\documentclass[
	a4paper, 
	10pt, 
	unnumberedsections, 
	twoside, 
]{LTJournalArticle}

\usepackage{times}  
\usepackage{helvet}  
\usepackage{courier}  
\usepackage{graphicx} 
\usepackage{caption} 
\usepackage{subcaption}
\usepackage{algorithm}
\usepackage{algorithmic}

\usepackage{amssymb,amsfonts,amsthm,bm,dsfont,amsmath}
 \usepackage{array}
 \newcolumntype{P}[1]{>{\centering\arraybackslash}p{#1}}
 \newtheorem{theorem}{Theorem}
\newtheorem{remark}{Remark}

\newtheorem{proposition}{Proposition}

\newtheorem{lemma}{Lemma}

\usepackage{comment}
\usepackage{newfloat}
\usepackage{listings}
\DeclareCaptionStyle{ruled}{labelfont=normalfont,labelsep=colon,strut=off} 
\floatstyle{ruled}
\newfloat{listing}{tb}{lst}{}
\floatname{listing}{Listing}

\title{Federated Aggregation of Mallows Rankings: \\ A Comparative Analysis of Borda and Lehmer Coding} 

\author{%
	Jin Sima\textsuperscript{1}, Vishal Rana\textsuperscript{1} and Olgica Milenkovic\textsuperscript{1}\thanks{Corresponding author: milenkov@illinois.edu\\ }
}

\date{\footnotesize\textsuperscript{\textbf{1}}Department of Electrical and Computer Engineering, University of Illinois Urbana-Champaign\\ 
}

\renewcommand{\maketitlehookd}{%
	\begin{abstract}
		\noindent Rank aggregation (RA) is the process of merging different rank-ordered lists into a single, representative consensus ranking. In certain settings, such as medical and biological data sharing and analysis, the rankings are stored at physically distributed locations and have to be kept private. This motivates the study of \emph{federated rank aggregation protocols,} which allow for distributed, private and communication-efficient learning across multiple clients with local data samples. Here we describe the first known collection of approaches for federated rank aggregation that adapt Borda scoring and Lehmer codes. 
        Our focus is on establishing sample complexity results for federated algorithms on Mallows distributions with known scaling factor $\phi$ and unknown centroid permutation $\sigma_0$.

        The federated Borda approach is a combination of local client scoring and nontrivial quantization of the scores, coupled with privacy-preserving protocols. The results reveal that for $\phi \in [0,1)$, and arbitrary $\sigma_0$ of length $N$, it suffices for each of the $L$ clients to locally aggregate $\max\{C_1(\phi), C_2(\phi)\frac{1}{L}\log \frac{N}{\delta}\}$ rankings, where $C_1(\phi)$ and $C_2(\phi)$ are constants that depend on $\phi$, quantize the result, and send it to the server who can then recover $\sigma_0$ with probability $\geq 1-\delta$. The total communication complexity scales as $NL\log N$. To the best of our knowledge, these results constitute the first rigorous analysis of Borda's method for both the centralized and distributed setting under the Mallows model. 
        
        The idea behind the federated Lehmer coding approach is to create a local Lehmer code for each client ranking dataset, by using a coordinate-majority aggregation approach. Since the server cannot find the majority of local coordinates in an efficient and privacy-preserving manner, we once again resort to specialized quantization techniques. The results reveal that for $\phi$ satisfying $\phi+\phi^2<1+\phi^N$, and arbitrary $\sigma_0$ of length $N$, it also suffices for each of the $L$ clients to locally aggregate $\max\{C_3(\phi), C_4(\phi)\frac{1}{L}\log \frac{N}{\delta}\}$ rankings, where $C_3(\phi)$ and $C_4(\phi)$ are constants that depend on $\phi$. The clients communicate truncated histograms of the Lehmer coordinate to the server who can then recover $\sigma_0$ with probability $\geq 1-\delta$. The communication complexity of the method is $\sim O(N\log N L\log L)$. 
        
        We also report findings from a comparative study of our federated rank aggregation approaches on synthetic and real-world data, such as Sushi preferences, Jester, two prioritized Landmark gene expression datasets and votes from the Primary Maine Governor Election 2018. 
	\end{abstract}
}

\begin{document}

\maketitle 


\section{Introduction}
Federated learning (FL) is a distributed data processing technique that enables training models across decentralized devices (clients) in possession of private local datasets that cannot be shared~\cite{konevcny2016federatedlearn,konevcny2016federatedopt,kairouz2021advances}. 
In FL systems, clients securely transmit locally trained models to the server, while the server learns a global model by aggregating local models. The server is not allowed to infer information about local datasets beyond what is needed for oblivious global aggregation.

One challenge in designing FL methods is to deal with the heterogeneity of client datasets~\cite{zhao2018federated}.
When heterogeneity is such that the local datasets inherently exist at widely different scales, it is advisable to convert the data into rankings~\cite{marden1996analyzing}. Rankings are also naturally generated by voting, gene prioritization and recommendation systems, all of which require data privacy. In such settings, the most frequently pursued overall learning goal is to aggregate the local data and/or learn the local statistical ranking models. These  tasks can be addressed via \emph{rank aggregation methods}~\cite{ailon2008aggregating,awasthi2014learning,coppersmith2006ordering,chen2009toppgene,kim2015hydra}. 

The above considerations motivate the study of \emph{federated rank aggregation} (FRA), where the goal is to aggregate local, private ranking information at the server without accessing data at the local client level. Such a distributed learning setting is especially important for \emph{multiomics and voting data,} which are subject to stringent privacy constraints and inherently distributed across different geographic regions. 
Despite its utility, FRA has not been previously considered in the machine learning literature.

Even in the centralized setting, the rank aggregation problem may be challenging, depending on the aggregation objective. Under the sum-of-distances-from-the-consensus objective, it is known to be NP-hard~\cite{dwork2001rank,ailon2008aggregating,coppersmith2006ordering}. Although many straightforward constant-factor approximation approaches for finding the solution to the problem are known~\cite{ailon2008aggregating}, existing theoretical analyses mostly focus on improving the approximation results for general collections of rankings. Some other frequently used aggregation algorithms, such as Borda scoring~\cite{lansdowne1996applying}, do not rely on a specific objective function. 

Less is known regarding rigorous \emph{sample complexity} studies for aggregating rankings generated by statistical models such as the Mallows~\cite{mallows1957non,ceberio2015review,busa2019optimal} and related models~\cite{kateri2022generalized,feng2022mallows} via specific aggregation algorithms. Some works~\cite{meila2012consensus,lu2011learning} consider estimating the model parameters of Mallows-like samples~\cite{vitelli} but either do not focus on general rank aggregation methods and do not address computational issues or only work with pairwise preferences. Other lines of work~\cite{li2017efficient} propose approaches for rank aggregation of Mallows and ''partial'' Mallows samples based on \emph{Lehmer codes}, which enable parallel aggregation and lead to a tractable problem analysis (see also the Related Works section). 

For the FRA problem, we focus on the statistical Mallows model, and derive new results for the client sample complexities of Borda and Lehmer code federated aggregation methods~\cite{lansdowne1996applying}. Importantly, we also describe and analyze new compression/quantization methods that enable secure aggregation and ensure low communication complexity. Our contributions are as follows: 

\textbf{1.} We introduce the problem of Federated Rank Aggregation (FRA) in which the clients are assumed to hold private rankings data insufficient to generate an accurate local consensus, but are able to share their local consensus in a communication-efficient and privacy-preserving manner with the server who accurately reconstructs the consensus.

\textbf{2.} We propose two new FRA algorithms and rigorously analyze their client sample complexity and server performance under the Mallows model. The methods include creating local consensus ranking by encoding the information in the rankings into Borda scores and  Lehmer encodings, quantizing the local aggregates using statistical information about the coordinates of the aggregates and exploiting the concentration properties of the coordinates and sparsity of encodings.
Our sample complexity analysis for the Borda aggregation method on Mallows models is the first of its kind for both the centralized and distributed setting.


\textbf{3.} We 
provide lower bounds on the communication complexity needed for FRA with secure aggregation features. The communication complexity bounds are based on local quantization methods that take into account that the server needs to aggregate large score values (Borda) or use majority approaches (Lehmer) that are hard to combine with standard secure aggregation protocols.

\textbf{5.} We present numerical results for both synthetic datasets generated via Mallows models with different scales and centers, and real datasets: \emph{Sushi} (food preferences), \emph{Jester} (joke preferences) ~\cite{kamishima2003nantonac,goldberg2001eigentaste}, and two prioritized cancer gene expression datasets from The Cancer Genome Atlas (TCGA), pertaining to subsets of Landmark genes. 
Our results indicate that FRA via Borda scoring exhibits excellent reconstruction accuracy and communication complexity on all examined data. 

The paper is organized as follows. The Related Works Section reviews the literature in the field, while the notation and key concepts are outlined in the Preliminaries Section. The Main Results Section contains our analytical findings pertaining to FRA via Borda and Lehmer codes. The section also includes a description of the aggregation, quantization and secure transmission methods and reports bounds on the sample and communication complexity of the FRA strategies. The Experiments Section presents our supporting simulation results for synthetic and real-world datasets.

\section{Related Works} \label{sec:relatedworks}
\textbf{Rank aggregation.} Rank aggregation is a method for constructing the consensus of a set of rankings~\cite{dwork2001rank} and it has been extensively studied in machine learning~\cite{chakraborty2022fair,klementiev2007unsupervised,korba2017learning,liang2018manifold}, theoretical computer science~\cite{dinu2006efficient,ailon2008aggregating,alabi2022private}, voting and recommendation systems~\cite{baltrunas2010group,schalekamp2009rank,balchanowski2023comparative}, as well as in computational biology~\cite{chen2009toppgene,kolde2012robust}. Often, the problem formulation requires one to minimize the total distance between a consensus and the data rankings~\cite{lin2010rank}, where the distance is either the Kendall $\tau$, the Spearman footrule or some other distance measure between rankings~\cite{diaconis1977spearman,kumar2010generalized}. For the Kendall $\tau$ distance, this form of rank aggregation is NP-hard~\cite{bartholdi1989voting}. As a result, many approximate algorithms are proposed for the problem, ranging from the simplest approach that suggests to pick a random permutation from the set (PICK-A-PERM), ``best of FAS-PIVOT and PICK-A-PERM'' algorithm~\cite{ailon2008aggregating} to the Spearman footrule matching method~\cite{ailon2008aggregating,alvin2023approximate}. Other approximation algorithms that can be used in general settings are Borda's and other score-based aggregation methods~\cite{lu2011learning}. There also exist empirical studies comparing various rank aggregation methods on real-world data, including~\cite{balchanowski2023comparative}. Nevertheless, no methods for FRA are currently known.

\textbf{The Mallows model.} The Mallows model is a frequently used probabilistic model for ranking data which assumes that the probability of sampling a ranking decays exponentially with its Kendall-$\tau$ distance to a centroid ranking. One key estimation problem for the Mallows model is to determine the sample complexity for recovering the centroid (since given the centroid, the scaling parameter can be recovered using convex optimization methods~\cite{fligner1988multistage}). Another algorithm for recovering the centroid under the Mallows models was suggested in~\cite{li2017efficient}, which uses Lehmer codes suitable for parallel aggregation. The algorithm only guarantees recovery for a constrained set of scaling parameter values. Recent focus in the field has been on estimating mixtures of Mallows models~\cite{awasthi2014learning,chierichetti2015learning}. 

\textbf{Federated methods for learning to rank.}  Prior works have so far only focused on the problems of federated learning to rank and attacks on learning to rank methods~\cite{mozaffari2021frl,kharitonov2019federated,wang2021efficient}, and federated online pairwise comparison ranking~\cite{lin2024federated}. Learning to rank operates by training models to rank results for different queries, while federated counterparts assume in addition that the training data is distributed and private. FRA, on the other hand, has the very different goal of learning a ranking that is a consensus of distributed ranking datasets. It also involves compressing local ranking data for efficient communication and ensuring that data privacy is maintained via secure aggregation. In the context of data privacy, a related line of work is on \emph{cetralized} differentially private rank aggregation~\cite{hay2017differentially}. 

\section{Preliminaries} \label{sec:preliminaries}
Let $[N]=\{1,\ldots,N\}$. A (complete) ranking (alternatively referred to as a permutation) $\sigma=(\sigma(1),\ldots,\sigma(N))$ of length $N$ is a bijection $\sigma: [N] \to [N]$, where $\sigma(i)=j$, $i,j\in [N],$ indicates that element $i$ is ranked at position $j$. The highest ranked element has position one. Also, $\mathbb{S}_N$ stands for the set of all permutations $\sigma$ of length $N$, i.e., the symmetric group of order $N!$. 

\textbf{Rank Aggregation approaches.} One family of aggregation methods relies on minimizing the total distance of the ranking dataset from a aggregate (consensus) ranking. There are several different distances and objectives used in practice. The most common distance is the \emph{Kendall $\tau$} distance, which for $\sigma_1,\sigma_2\in \mathbb{S}_N$ is defined as
\begin{align*}
K_\tau(\sigma_1,\sigma_2)=|\{(i,j): \,&\sigma_1(i)>\sigma_2(i) \textup{ and }\\
&\sigma_1(j)<\sigma_2(j),i,j\in[N]\}|.    \end{align*}
In words, the Kendall $\tau$ distance counts the number of pairs of elements in inverted order within the two permutations, i.e., the number of \emph{inversions}. A rank aggregation objective is the cumulative Kendall $\tau$ distance between the aggregated ranking $\sigma^*$ and data rankings $\sigma_m\in \mathbb{S}_N$, $m\in[M]$,
$$\sigma^*=\arg\min_{\sigma\in \mathbb{S}_N}\sum^M_{m=1}K_\tau(\sigma^*,\sigma_m).$$
The ranking $\sigma^*$ is referred to as the \emph{Kemeny consensus.} 
Algorithms such as FAS-PIVOT \cite{ailon2008aggregating} and Spearman footrule approximation ~\cite{dwork2001rank} aim to optimize cumulative distance measures. 

Another family of rank aggregation methods does not rely on specific objective functions. Among them, a version of Borda's method \cite{lansdowne1996applying} stands out due to its simplicity and efficiency. It takes coordinate-wise averages of all rankings in the dataset and ranks the averages to generate the consensus. Besides Borda, Lehmer encoding-based algorithms~\cite{li2017efficient}, which convert rankings in the dataset into corresponding Lehmer codes, and then form a consensus via coordinate-wise majority of the Lehmer encodings, also do not involve objective functions. 
For a permutation $\sigma$, its Lehmer encoding vector $\mathcal{L}_{\sigma}$ is defined as $\mathcal{L}_{\sigma}(i)=|\{t\in[i-1]:\sigma(t)>\sigma(i)\}|,$ $i \in [N]$. A Lehmer encoding $\mathcal{L}_{\sigma}$ and its inverse 
$\mathcal{L}^{-1}(\mathcal{L}_{\sigma})=\sigma$ 
can be efficiently computed with time complexity $O(N)$~(see the references in~\cite{li2017efficient}).
 
\textbf{The Mallows Model.} The Mallows model is a probabilistic ranking model that assumes that data rankings $\sigma_1,\ldots,\sigma_M$ are generated in an iid manner,  according to a distribution parameterized by a centroid permutation $\sigma_0\in \mathbb{S}_N$ and a scaling factor $\phi \in (0,1)$ such that  
\begin{align}\label{eq:mallow}
   \sigma \sim \frac{\phi^{K_\tau(\sigma_0,\sigma)}}{Z},
\end{align}
where $Z=\sum_{\sigma'\in \mathbb{S}_N}\phi^{K_\tau(\sigma_0,\sigma')}$ is a normalization constant that can be shown to equal $\prod^N_{i=1}(\sum^i_{j=0}\phi^j)$ for any $\sigma_0\in \mathbb{S}_N$. 
One important fact about the Mallows model is that the coordinates of the vector $f_{\sigma_0,\sigma}$,
$$f_{\sigma_0,\sigma}(i)=|\{t\in[i-1]:\sigma(\sigma^{-1}_0(t))>\sigma(\sigma^{-1}_0(i))\}|,$$ where $\sigma^{-1}_0$ is the inverse of $\sigma_0$, are independent and follow truncated geometric distributions, i.e.,
\begin{align}\label{eq:truncatedgeo}
    Pr(f_{\sigma_0,\sigma}(i)=j)= \frac{\phi^j}{\sum^{i-1}_{j=0}\phi^j},~j\in[i-1],i\in[N]. 
\end{align}
Note that $f_{\sigma_0,\sigma}=\mathcal{L}_{\sigma}$ equals the Lehmer code when $\sigma_0 = e,$ i.e., when the centroid equals the identity permutation. 

It is known that the maximum likelihood estimator (MLE) of the centroid of the Mallows model, $\sigma_0,$ has the form of the Kemeny aggregate~\cite{meila2012consensus}. Hence, given that the general Kemeny aggregate problem is NP-hard, we focus on analyzing the performance of FRA algorithms on the Mallows model, where the focus is on recovering (estimating) $\sigma_0$ in a distributed and privacy-preserving manner.




\section{Main Results: FRA for the Mallows Model} \label{sec:mainresults}
The underlying assumption behind our analyses are as follows. A collection of $L$ ranking datasets, $\sigma_{\ell,1},\ldots,\sigma_{\ell,m_\ell}$, $\ell\in [L]$, is generated via iid sampling from a Mallows distribution, according to~\eqref{eq:mallow}. These datasets are in possession of $L$ clients, such that $\sigma_{\ell,1},\ldots,\sigma_{\ell,m_\ell}$ is the private dataset at  client $\ell\in[L],$ comprising $m_{\ell}$ samples. The number of clients $L\geq 2$ that can be accommodated depends on the aggregation method. Note that due to the iid sampling procedure (i.e., sampling with replacement) at each client, it is possible for different clients to share the same ranking or one client to contain repeated rankings. The repeat frequencies depend on $\phi$.

It is also assumed that $\phi$ in the Mallows model is known to both the clients and the server, while $\sigma_0$ is unknown. The knowledge of $\phi$ is only needed for deriving rigorous sample complexity results, but is not used in the Borda and Lehmer code aggregation algorithms themselves (except for the final quantization steps which are fairly robust to the choice of $\phi$). The goal is to learn the Mallows centroid permutation $\sigma_0,$ i.e., find a Kemeny aggregate at the server through a single round of client-to-server communication. Generally, the goal is to find a Kemeny ranking for arbitrary client datasets: in this case, the Borda and Lehmer code methods still apply but do not come with provable performance guarantees.

In FL settings, each client has to protect the privacy of its ranking without revealing any information (such as the number of rankings available, their identity, statistics etc) to the server. This is achieved through secure aggregation~\cite{bonawitz2017practical}, where each client $\ell\in[L]$ trains a local model based on its local permutation data $\sigma_{\ell,m}$, $m\in[m_\ell]$ and encodes the local model into a message $y_\ell$ whose length depends on the aggregation scheme used. A ``noisy'' version $y_\ell+z_\ell$ of the message is sent to the server so that the local model at client $\ell$ is obfuscated. The server adds the noisy encodings of the local models $\sum^L_{\ell=1}y_\ell+\sum^L_{\ell=1}z_\ell$. The samples $z_\ell$, $\ell\in[L],$ are required to satisfy $\sum^L_{\ell=1}z_\ell=0$ so that the aggregate at the server reduces to $\sum^L_{\ell}y_\ell$.  In addition, it is desirable to minimize the communication cost, which is the total number of bits needed to represent all messages $y_{\ell}+z_\ell$, $\ell\in[L]$.

\begin{figure}[t]
    \centering
    \includegraphics[width=\linewidth]{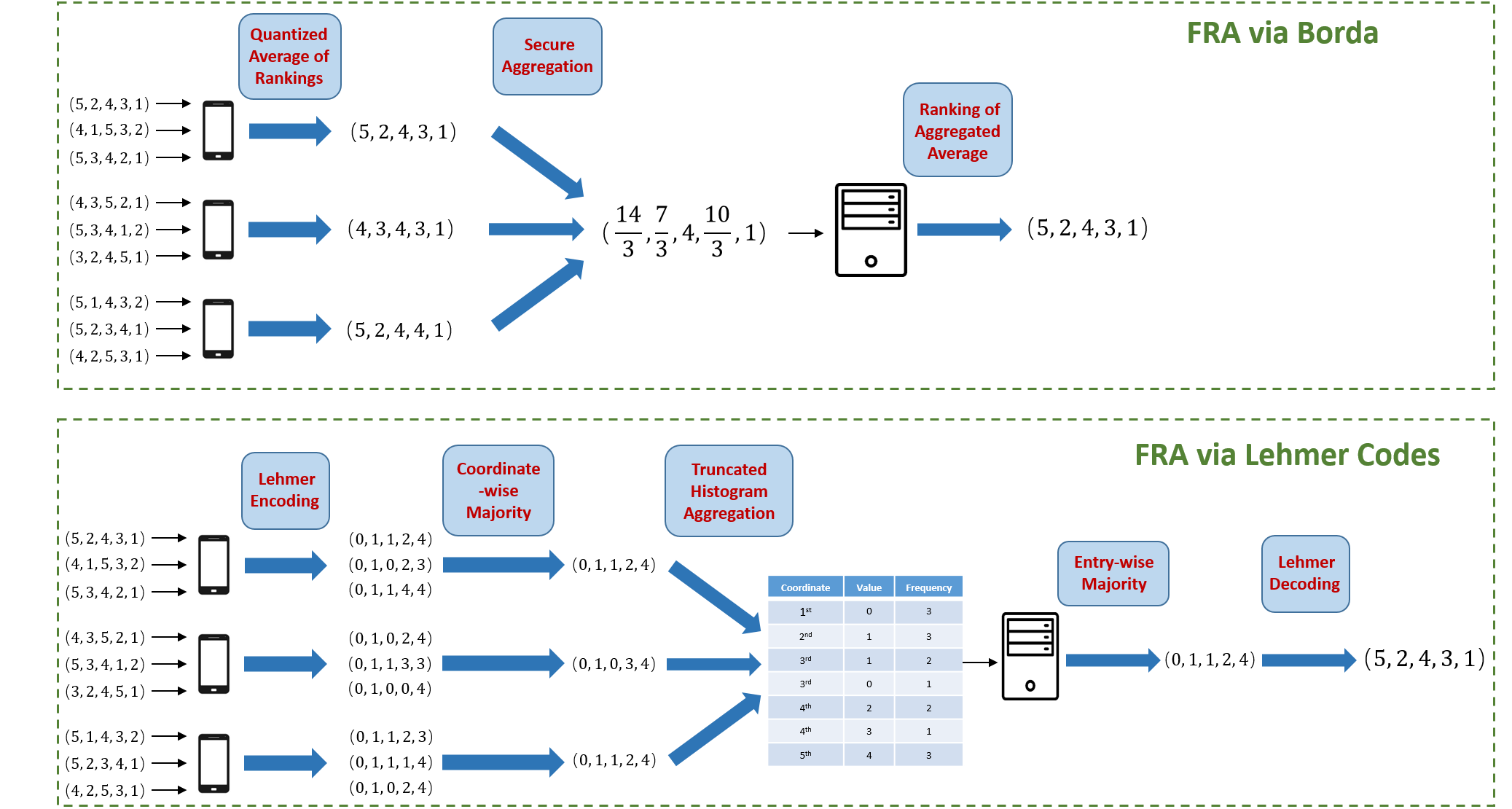}
    \caption{Overview of our proposed FRA algorithms based on normalized Borda scores (top) and Lehmer codes (bottom). In Borda's algorithm, each client computes the average of local permutations and nonuniformly quantizes their values to arrive at a vector representation that is not necessarily a permutation. The server aggregates the quantized messages into a ranking. In the algorithm based on Lehmer codes, each client computes the Lehmer encoding of each local permutation and their entry-wise majority. Then, each client encodes the truncated histogram of the entry-wise majority of the Lehmer codes, guided by concentration results. The server computes the overall entry-wise majority of the client codes and performs Lehmer decoding.}
    \label{fig:flowchart}
\end{figure}

Next, we describe our FRA methods, whose flowcharts are depicted in Figure~\ref{fig:flowchart}.

\subsection{FRA of Mallows Samples via Borda's Method}
We start with an analysis of the centralized Mallows rank aggregation algorithm based on Borda scoring (Algorithm~\ref{alg:average}). Despite being a simple and widespread approach that also lends itself to distributed implementations, it was not previously analyzed. The algorithm recovers $\sigma_0$ with probability at least $1-\delta$ whenever the total number of samples $M\ge O(\log \frac{N}{\delta}),$ for any fixed $0<\phi<1$. 

Since sums of sums of scores equal to the global sums, it is straightforward to adapt this algorithm to the FRA setting, while maintaining the same total sample complexity $\sum^L_{\ell=1}m_\ell=O(\log \frac{N}{\delta}).$ The only complication arises with respect to the communication cost, since the sums may be large integer values: to mitigate the problem, we normalize and quantize the data so as to minimize the distortion in the overall ranking scores. As a result, our FRA algorithm requires  $\max\{O(\frac{\log\frac{N}{\delta}}{L}),$ $O(1)\}$ ranking samples at each client, maintaining the total sample complexity of the centralized method,  $\sum^L_{\ell=1}m_\ell=O(\log \frac{N}{\delta}),$ provided that $L\le O(\frac{N}{\delta})$. The findings above rely on some known and some newly derived properties of the Mallows distribution and the Borda scores of the rankings (e.g., Lemma 1 and 2, whose proofs are delegated to the Supplement). 

\begin{algorithm}[H]
 \caption{Centralized RA via Borda Scoring}
   \label{alg:average}
   \begin{algorithmic}[1]
    \STATE \textbf{input:} Permutations $\sigma_1,\ldots,\sigma_M$ sampled in an iid manner from the Mallows distribution~\eqref{eq:mallow}.
    \STATE Compute the average $A_i=\frac{\sum^M_{m=1}\sigma_m(i)}{M}$ and reorder the $A_i$'s, $i\in [N]$, so that $A_{i_1}<A_{i_2}<\ldots<A_{i_N}$. Let $\hat{\sigma}_0\in \mathbb{S}_N$ be such that $\hat{\sigma}_0(j)=i_j$ for $j\in[N]$. 
    \RETURN An estimate of the centroid $\hat{\sigma}_0$.
  \end{algorithmic}
\end{algorithm}

The first result shows that the value of the $i$th coordinate of each permutation $\sigma$ generated according to a Mallows distribution is dominated by a two-sided geometric distribution.
\begin{lemma}\label{lem:geometric}
Let $\sigma\in \mathbb{S}_N$ be a permutation generated according to Mallows distribution \eqref{eq:mallow}. Then,
\begin{align*}
Pr(\sigma(i)=j)\le \phi^{|j-\sigma_0(i)|}, \, i,j\in [N].  
\end{align*} 
\end{lemma}
Lemma \ref{lem:geometric} guarantees that large deviations of $\sigma(i)$ from $\sigma_0(i)$ are unlikely. Another claim is that the the expectations $E[\sigma(i)]$, $i\in[N],$ ``follow'' the same order as $\sigma_0$.
\begin{lemma}\label{lem:exporder}
Let $\sigma$ be a permutation drawn from the Mallows distribution ~\eqref{eq:mallow}. Then, $E[\sigma(i)]>E[\sigma(j)]$ iff $\sigma_0(i)>\sigma_0(j)$.
\end{lemma}
Based on Lemmas ~\ref{lem:geometric} and~\ref{lem:exporder}, we can establish the following result for centralized Borda aggregation.
\begin{theorem}\label{thm:averaging}
Let $\sigma_1,\ldots,\sigma_M$ be a set of permutations independently sampled from a Mallows distribution~\eqref{eq:mallow} with $0<\phi<1.$ If  
\begin{align}\label{eq:lowerboundM}
    M\ge \frac{16C^2(1+\phi)^2\log \frac{2N}{\delta}}{(1-\phi)^2}, \text{ where}
\end{align}
\begin{align}\label{eq:C}
C= \max\{ \frac{8\ln \frac{128(1+\phi)^2}{(1-\phi)^2(1-\sqrt{\phi})}}{\ln \frac{1}{\phi}}, e^{\frac{\ln \frac{1}{\phi}}{16}},\sqrt{\frac{2(1-\phi)}{(1+\phi)\ln \frac{1}{\phi}}} \},    
\end{align}
then Algorithm~\ref{alg:average} returns $\sigma_0$ with probability at least $1-\delta$.    
\end{theorem}

\begin{figure}[htb]
    \centering
\includegraphics[width=\linewidth]{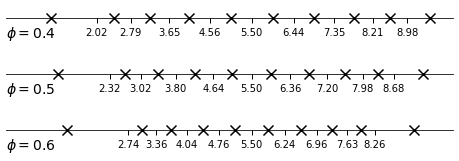}
    \caption{Quantization bins (bars) and centroids (crosses) for Borda aggregates, and for different values of $\phi$, computed via Monte Carlo methods with $100,000$ samples.}
    \label{fig:bin_size}
\end{figure}

FRA via Borda's algorithm entails of each client to compute the coordinate-wise sums of scores of its local rankings and transmits it to the server through a secure aggregation scheme ~\cite{bonawitz2017practical}. The server then averages the local sums (and thus obtains the coordinate-wise sum of the collection of all rankings), which leads to the same result as in the centralized setting. However, such a scheme results in an unnecessary high communication cost whenever some client has many rankings to locally aggregate over. If the scores are normalized to averages, which are significantly smaller than the sums, then the server needs to know the total number of samples per client in order to compute the average. To maintain privacy regarding the number of samples per client, we propose a FRA Borda algorithm where each client quantizes the average scores of the ranking coordinates, and then transmits them to the server (Algorithm~\ref{alg:fedaverage}). The quantization rule is based on the claim from Lemma~\ref{lem:exporder} that $E[\sigma(\sigma^{-1}_0(1)]<\ldots<E[\sigma(\sigma^{-1}_0(N)]$. The values $E[\sigma(\sigma^{-1}_0(i)]$, $i\in[N]$, are independent of the choice of $\sigma_0$. Thus, $E_i=E[\sigma(\sigma^{-1}_0(i)]$ can be used as the $i$th quantization centroid for the aggregates, and as outlined in the Supplement, it can be efficiently computed using Monte Carlo methods or simple recursions.  
Quantization centroids and threshold for different values of $\phi$ are given in Figure~\ref{fig:bin_size}.
\begin{theorem}\label{thm:fedaverage}
Let $\sigma_{\ell,m}$, $m\in[m_\ell]$, be a set of iid samples from the Mallows distribution~\eqref{eq:mallow} distributed over $\ell\in[L]$ clients. Let $C$ be given by~\eqref{eq:C}. Then, if for all $\ell\in[L]$ 
\begin{align*}
m_\ell \ge \max\{&\frac{256(1+\phi)^2}{(1-\phi)^2},\\&\frac{C^2(260\ln(\frac{2N}{\delta}+\ln(\frac{4}{3}L))(1+\phi)^2}{L(1-\phi)^2}\},    
\end{align*} 
Algorithm~\ref{alg:fedaverage} returns $\sigma_0$ with probability at least $1-\delta$. The communication complexity scales as $NL\log N$.
\end{theorem}
\subsection{FRA of Mallows Samples via Lehmer Codes}
Our second FRA algorithm for estimating $\sigma_0$ is based on the coordinate-wise majority of Lehmer codes of individual rankings, which requires $\max\{O(\frac{\log\frac{N}{\delta}}{L}),O(1)\}$ samples at each client and ensures a total communication cost $O(LN\max\{\log N,\log M\}\log L),$ provided that $\phi+\phi^2<1+\phi^N$, where $M=\sum^L_{\ell=1}m_\ell.$ 

The key steps of the Lehmer FRA approach are listed in Algorithm~\ref{alg:lehmerfed}. They include the idea from~\cite{li2017efficient} to convert rankings into Lehmer codes and then use the \emph{majority value} of each coordinate for the consensus Lehmer code, with the result summarized below. 

\begin{theorem}\label{lem:Pantheorem}
    \cite{li2017efficient} Let $\sigma_1,\ldots,\sigma_M$ be a set of permutations independently generated by the Mallows distribution~\eqref{eq:mallow}. If $\phi+\phi^2<1+\phi^N$ and $M\ge \frac{2(1+p)^2}{(1-p)^4}\log \frac{N^2}{2\delta }$, where $p=\frac{\sum^{N-1}_{u=1}\phi^u}{1+\sum^{N}_{u=3}\phi^u}$, then the coordinate-wise majority of $\mathcal{L}_{\sigma_m}$, $m\in[M]$ equals $\mathcal{L}_{\sigma_0}$ with probability at least $1-\delta$.
\end{theorem}
In the federated setting, a client may not have sufficiently many local rankings to recover the centroid permutation $\sigma_0$. Hence, Theorem~\ref{lem:Pantheorem} cannot be applied directly. Furthermore, one has to consider how to transmit the local Lehmer majority aggregates in a secure and efficient manner so that the \emph{server} can perform \emph{one more round of majority-voting (rather than averaging)} of the local client Lehmer codes.  

\begin{algorithm}[H]
 \caption{FRA via Borda's Method}
   \label{alg:fedaverage}
   \begin{algorithmic}[1]
    \STATE \textbf{input:} Permutations $\{\sigma_{\ell,1},\ldots,\sigma_{\ell,m_\ell}\}_{\ell\in[L]}$ sampled from the Mallows distribution \eqref{eq:mallow}, stored at $L$ clients. Client $\ell\in[L]$ holds permutations $\{\sigma_{\ell,1},\ldots,\sigma_{\ell,m_\ell}\}$.
    \STATE Each client $\ell$ computes the coordinate-wise average of its local permutations $A_{\ell}(i)=\frac{\sum^{m_\ell}_{m=1}\sigma_m(i)}{m_\ell},$ and makes an estimate $\hat{\sigma}_{\ell}\in \mathbb{S}_N$ of $\sigma_0$ such that $\hat{\sigma}_\ell(i)=\arg\min_{j\in [N]}|A_{\ell}-E_j|$. Note that $\hat{\sigma}_\ell$ can have repeated entries, i.e.,  $\hat{\sigma}_\ell(i_1)=\hat{\sigma}_\ell(i_2)$ for $i_1\ne i_2$. Each client $\ell$ sends $\hat{\sigma}_\ell$ using standard secure aggregation.
    \STATE The server aggregates $\hat{\sigma}_\ell$ and computes $A(i)=\frac{\sum^{L}_{\ell=1}\hat{\sigma}_\ell(i)}{L}$. Then the server reorders the values $A(i)$ in increasing order, $A(i_1)<A(i_2)<\ldots<A(i_N),$ and estimates $\hat{\sigma}_0(j)=i_j,$ for $j\in[N]$. 
    \RETURN An estimate of the centroid, $\hat{\sigma}_0,$ at the server.
  \end{algorithmic}
\end{algorithm}


 Avoiding majority computations and resorting to averaging Lehmer codes instead produces poor results~\cite{li2017efficient}. This leads to nontrivial issues that can be resolved by compressing the coordinate-wise majority of Lehmer codes, given that the coordinate values concentrate around the centroid coordinate, which leads to a sparse histogram of the Lehmer encoding of local rankings with high probability. Thus, the clients transmit the sparse Lehmer code histograms rather than the codes themselves. Clearly, histograms can be aggregated securely in a standard manner and \emph{no direct majority computations on the Lehmer codes} are needed at the server. Proving that the Lehmer code histograms of Mallows samples are sparse relies on a lemma from~\cite{li2017efficient} (stated in Proposition 1 in the Supplement), which can be used to show that the probability distribution of $i-\mathcal{L}_{\sigma_{\ell,m}}(i)$, $i\in[N]$, $\ell\in[L]$, centers at the value of $i-\mathcal{L}_{\sigma_0}(i)$ and decays exponentially. Hence, each client truncates the value of $\mathcal{L}_{\sigma_{\ell,m}}(i),$ for $i\in[N]$ and $m\in[
m_\ell],$ to the last 
\begin{align}\label{eq:I}
I=\lceil\log\Big(2\frac{\log (\frac{MN^2}{\epsilon})}{\log (\frac{1}{p})}+1\Big)\rceil
\end{align}
bits of its binary representation. Our corresponding result is formally stated in the next lemma.
\begin{lemma}\label{lem:majoritylehmer}
Let $\sigma_{\ell,m}$, $m\in[m_\ell]$, be a set of permutations independently generated by Mallows distribution~\eqref{eq:mallow} and distributed over all $\ell\in[L]$ clients. If $\phi+\phi^2<1+\phi^N$,
and $m_\ell\ge\max\{\frac{4(1+p)^2\ln\Big(6\frac{\ln (\frac{1}{1-p})}{\ln (\frac{1}{p})}\Big)}{(1-p)^2}, \frac{8(1+p)^2}{(1-p)^2}(\ln 2+\frac{\ln(\frac{NL}{\delta})}{L})\} $, then with probability at least $1-\delta$, more than half of the values of $v_\ell(i)$ across the $L$ clients, described in 
Algorithm~\ref{alg:lehmerfed}, equal $\mathcal{L}_{\sigma_0}(i),$ $i\in[N]$. Therefore, the majority of $v_\ell(i)'$s equals $\mathcal{L}_{\sigma_0}(i),$ for $i\in[N],$ with probability at least $1-\delta$. 
\end{lemma} 
In Algorithm~\ref{alg:lehmerfed}, each client transmits the coordinate-wise majority value of the Lehmer coding of local rankings without sending its corresponding frequency. One may think that the frequencies may improve the estimate, but our experiments and analyses confirm this not to be the case. Also, since the coordinates of the Lehmer code are independent, with truncated geometric distributions~\eqref{eq:truncatedgeo} whenever $\sigma_0=e$, one may also consider using Golomb codes~\cite{chandra2001system} to losslessly encode the Lehmer code coordinates. Golomb codes are variable-length codes that optimally (in terms of coding rate) compress geometrically distributed random variables (see the Supplement). However, the variable-length makes it difficult to ensure secure aggregation. Moreover, Lehmer codes do not follow truncated geometric distributions for $\sigma_0\ne e$.

Our main summary result is stated below.
\begin{theorem}\label{thm:lehmercodes}
Let $\sigma_{\ell,m}$, $m\in[m_\ell], \ell\in[L]$, be a set of permutations independently generated by the Mallows distribution~\eqref{eq:mallow}, placed across $L$ clients. If $\phi+\phi^2<1+\phi^N$ and $m_\ell\ge\max\{\frac{4(1+p)^2\ln\Big(6\frac{\ln (\frac{1}{1-p})}{\ln (\frac{1}{p})}\Big)}{(1-p)^2}, \frac{8(1+p)^2}{(1-p)^2}(\ln 2+\frac{\ln(\frac{NL}{\delta})}{L})\} $, where $p=\frac{\sum^{N-1}_{u=1}\phi^u}{1+\sum^{N}_{u=3}\phi^u}$, then the server can recover $\sigma_0$ with probability at least $1-\epsilon-\delta$ using Algorithm ~\ref{alg:lehmerfed}. The total communication cost equals $L\Big(N(\log N- \log\Big(2\frac{\log (\frac{MN^2}{\epsilon})}{\log (\frac{1}{p})}+1\Big)+\Big(2\frac{\log (\frac{MN^2}{\epsilon})}{\log (\frac{1}{p})}+1\Big)\log L\Big)=O(LN\log L\max\{\log N,\log M\}))$, where $M=\sum^L_{\ell=1}m_\ell$ is the total number of local sample rankings.  
\end{theorem}

\begin{algorithm}[H]
 \caption{FRA via Lehmer Encoding}
   \label{alg:lehmerfed}
   \begin{algorithmic}[1]
    \STATE \textbf{input:} Collections of permutations $\sigma_{\ell,1},\ldots,\sigma_{\ell,m_\ell}, \ell\in[L],$ generated according to the Mallows distribution~\eqref{eq:mallow}.
    \FOR{each client $\ell\in [L]$}    
    \STATE Let $v_\ell(i)$, $i\in[N]$ be the coordinate-wise majority of $\mathcal{L}_{\sigma_{\ell,m}}(i)$ among all $m\in [m_\ell]$. When there are ties for the majority counts of  $\mathcal{L}_{\sigma_{\ell,m}}(i)$, choose one randomly.
    \STATE Let $v^1_{\ell}(i)$, $i\in[N]$, be the value of the first $\max\{\lceil\log i\rceil-I,0\}$ bits of the binary representation of $v_\ell(i)$, where $I$ is defined in \eqref{eq:I} and $v^1_{\ell}(i)=0$ if $\lceil\log i\rceil-I\le 0$. Let $x^1_{\ell}(i)=v^1_{\ell}(i)+z^1_{\ell}(i)\bmod 2^{\max\{\lceil\log i\rceil-I,0\}}$, where $z^1_{\ell}(i)$ is uniformly distributed over $ \{0,\ldots,M2^{\max\{\lceil\log i\rceil-I,0\}}-1\}$ such that $\sum^L_{\ell=1}z^1_{\ell}(i)\equiv 0\bmod 2^{\max\{\lceil\log i\rceil-I,0\}}$ for $i\in[N]$.
    
    \STATE Let $\boldsymbol{v}^2_{\ell,i}\in\{0,1\}^{2^I}$ be the one-hot encoding of the value of the last $I$ bits of $v_\ell(i)$, i.e., $\boldsymbol{v}^2_{\ell,i}(j)=$
    \begin{align*}
        &\begin{cases}
            1, &\mbox{ if the value of the last $I$ bits of $v_\ell(i)$ equals $j$;}\\
            0,&\mbox{otherwise,}
        \end{cases}
    \end{align*}
    where $\boldsymbol{v}^2_{\ell,i}(j)$ is the $j$th coordinate of $\boldsymbol{v}^2_{\ell,i}$, $j\in[2^I]$.
    Let $\boldsymbol{x}^2_{\ell,i}=\boldsymbol{v}^2_{\ell,i}+\boldsymbol{z}^2_{\ell,i}\bmod (L+1)$, where $\boldsymbol{z}^2_{\ell,i}$ is a length $2^I$ vector with its  $j$th entry, $j\in[2^I]$, $\boldsymbol{z}^2_{\ell,i}(j),$ uniformly distributed over $\{0,\ldots,L\},$ and such that $\sum^L_{\ell=1}\boldsymbol{z}^2_{\ell,i}(j)\equiv 0 \bmod (L+1)$.  
    
    \STATE Client $\ell$ transmits  $(x^1_\ell(1),\boldsymbol{x}^2_{\ell,1},\ldots,x^1_\ell(N),\boldsymbol{x}^2_{\ell,N})$ to the server.
    \ENDFOR
    \STATE The server receives $\sum^L_{\ell=1}(x^1_\ell(1),\boldsymbol{x}^2_{\ell,1},\ldots,x^1_\ell(N),\boldsymbol{x}^2_{\ell,N})$ and retrieves the histogram $\sum^L_{\ell=1}\boldsymbol{x}^2_{\ell,i}$, which allows for computing the majority value of the integers encoded by the last $I$ bits of the binary representation of $v_\ell(i),$ $\ell\in[L]$, $i\in [N]$. Let $Maj(i)$ be the majority of the value of the  last $I$ bits of $v_\ell(i)$ across  $\ell\in[L]$. Let $V(i)$, $i\in[N]$, be the closest integer to $\frac{\sum^L_{\ell=1}x^1_\ell(i)}{L}$. The value of $V(i)$ is the estimate of the first $\max\{\lceil\log i\rceil-I,0\}$ bits of the majority of $v_\ell(i)$.  
    \STATE The server estimates $\mathcal{L}_{\sigma_0}(i)$ using $\hat{\mathcal{L}}_{\sigma_0}(i)=2^{\max\{\lceil\log i\rceil-I,0\}}V(i)+Maj(i).$ Then, it computes the inverse of the Lehmer code $(\hat{\mathcal{L}}_{\sigma_0}(1),\ldots,\hat{\mathcal{L}}_{\sigma_0}(N))$ to obtain $\hat{\sigma}_0.$
    \RETURN The estimate $\hat{\sigma}_0$ at the server.
  \end{algorithmic}
\end{algorithm}
\section{Experiments} \label{sec:experiments}

\begin{figure*}[!htb]
    \centering
    \begin{subfigure}[t]{0.33\textwidth}
        \centering
        \includegraphics[width=\linewidth]{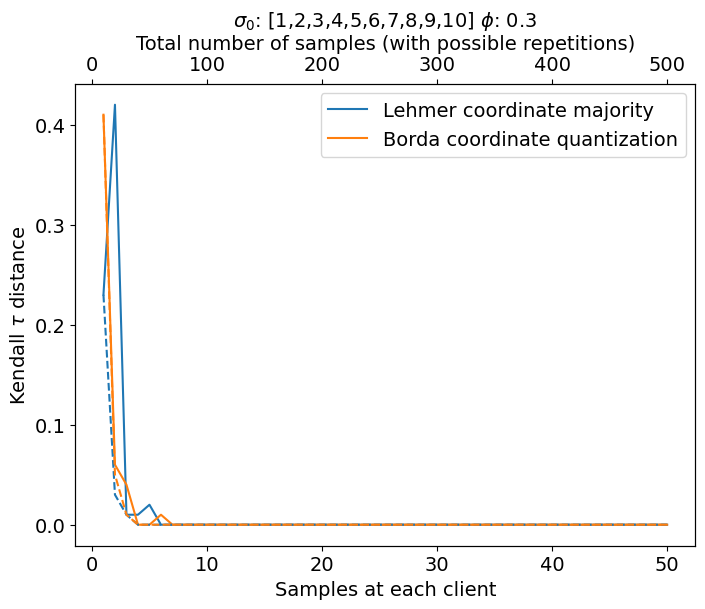} 
        \caption{$\sigma_0 = e, \phi=0.3$, $L=10$.} \label{}
    \end{subfigure}
    \hfill
    \begin{subfigure}[t]{0.33\textwidth}
        \centering
        \includegraphics[width=\linewidth]{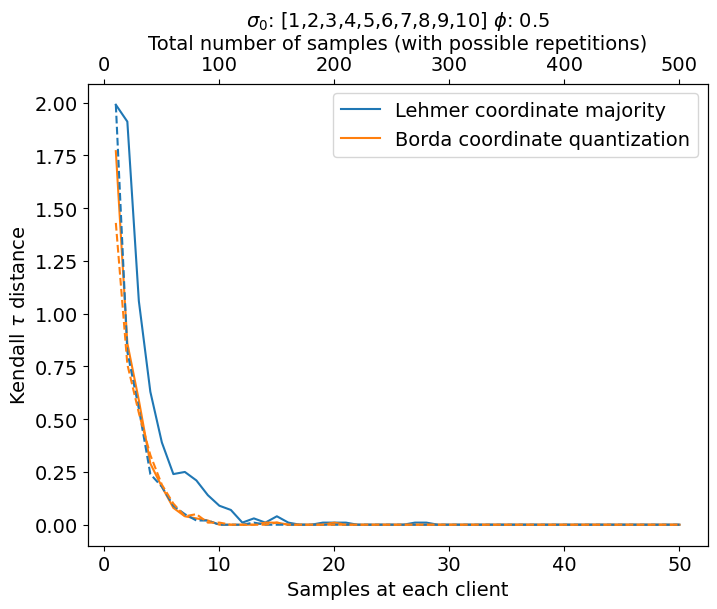} 
        \caption{$\sigma_0 = e, \phi=0.5$, $L=10$.} \label{}
    \end{subfigure}
    \begin{subfigure}[t]{0.33\textwidth}
        \centering
        \includegraphics[width=\linewidth]{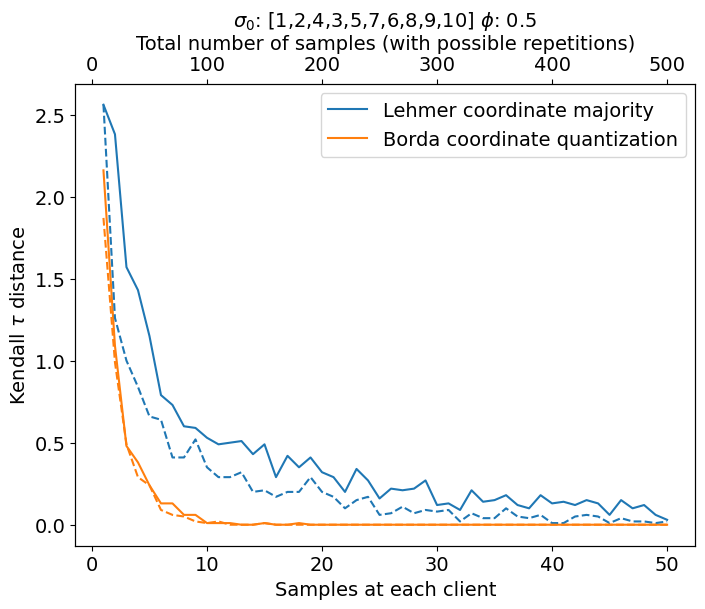} 
        \caption{$\sigma_0 \neq e, \phi=0.5$, $L=10$.} \label{}
    \end{subfigure}
    \hfill
    \begin{subfigure}[t]{0.33\textwidth}
        \centering
        \includegraphics[width=\linewidth]{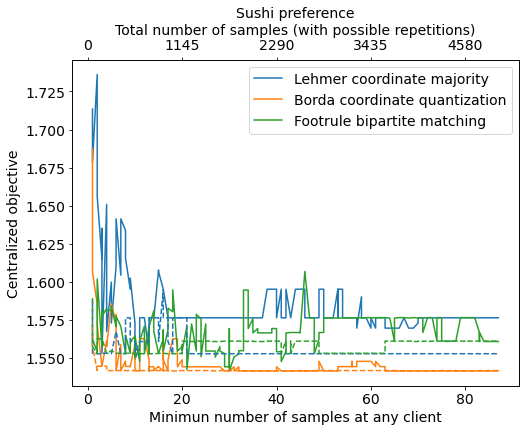} 
        \caption{$N=10, M=4,981$, $L=10$.} \label{fig:real_data_sushi}
    \end{subfigure}
    \hfill
    \begin{subfigure}[t]{0.33\textwidth}
        \centering
        \includegraphics[width=\linewidth]{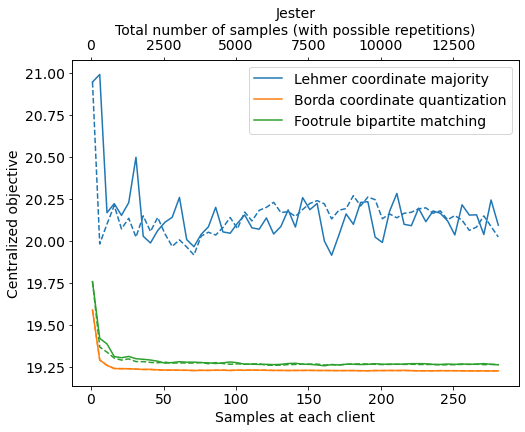} 
        \caption{$N=100, M=14,116$, $L=50$.} \label{fig:real_data_jester}
    \end{subfigure}
    \hfill
    \begin{subfigure}[t]{0.33\textwidth}
        \centering
        \includegraphics[width=\linewidth]{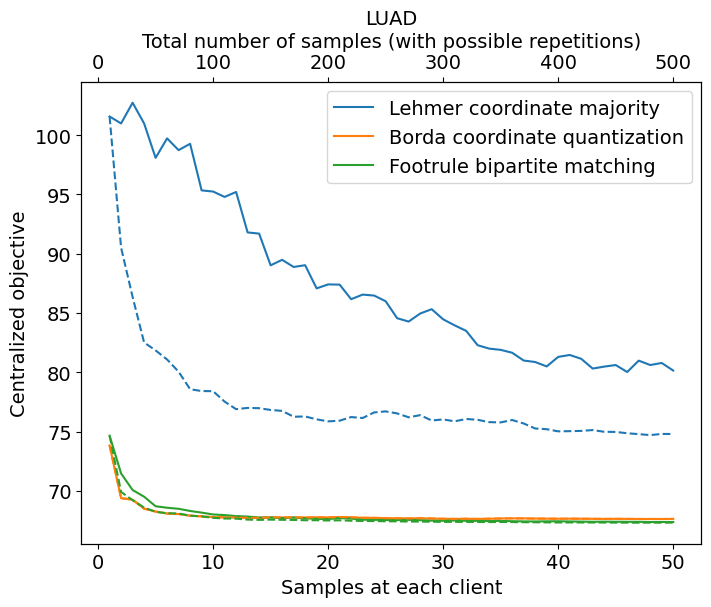} 
        \caption{$N=978, M=501$, $L=10$.} \label{fig:real_data_cancer}
    \end{subfigure}
    
    \caption{Comparison of the performance of various centralized and FRA methods (centralized results are depicted by dashed, while federated results are indicated by solid lines): (a-c) plots of the average Kendall $\tau$ distances between $\sigma_0$ and its estimate, for different values of $\phi$. (d-e) show the performance of our algorithms on Sushi preference, Jester, and prioritized cancer gene expression data, respectively. Here, the performance is evaluated using the Kemeny objective normalized by the permutation length. Results pertaining to the minimum weight bipartite matching aggregation algorithm are also shown~\cite{dwork2001rank}.}
    \label{fig:mallows_kendall}
\end{figure*}
We evaluate the performance of Borda and Lehmer FRA on synthetic Mallows datasets and real-world ranking data, and compare their performance with that of corresponding centralized versions of the algorithms. For real-world data, we also compare the performance of FRA methods with that of  minimum weight bipartite matching algorithms that use  Spearman's Footrule distance~\cite{dwork2001rank}.

\textbf{Synthetic Data.} We first examine the performance of our FRA methods on Mallows samples generated for different values of $\phi$ ~\cite{mallows1957non,collas2021concentric}. 

In the first set of experiments, we set  $L=10$. At each client $l\in [L]$, we generate a set of $m_l \in \{1,\cdots,50\}$ rankings with $\sigma_0=e$, and repeat the process for different values of $\phi \in (0,0.6]$ (since the Lehmer FRA method comes with provable performance guarantees only when $\phi \leq 0.6$). Each client and the server follow the algorithmic steps described in the previous section. The results are compared with those of the centralized protocols. The experiments are repeated for each set of parameter choices $100$ times. The reported results (Figure~\ref{fig:mallows_kendall}a and 3b) are the average Kendall $\tau$ distances between $\sigma_0$ and $\sigma^*$ as well as the average distance between $\sigma_0$ and the centralized estimate.  

 In this case, the Borda and Lehmer FRA methods, which share the same sample complexity, perform nearly identically and almost match the centralized aggregation benchmarks. The experiments are repeated for $\sigma_0 \neq e$ (Figure~\ref{fig:mallows_kendall}c).   
In this case, Borda significantly outperforms Lehmer FRA, as the coordinates of the Lehmer codes do no longer follow a truncated geometric distribution. More results are available in the Supplement. 

    

To evaluate the impact of the number of clients on the performance of the algorithms, we also performed another set of experiments where we fix the number of samples at each client $m_l = 10$ and increased the number of clients $L$ from $1$ to $50$. Again, we evaluate the performance for two different choices of centroids $\sigma_0$ of length $N=10$. We vary the values of $\phi \in (0,0.6)$ and repeat our experiments $100$ times. The results can be found in the Supplement.

\textbf{Real-World Data.} We evaluate the performance of FRA on Sushi preferences~\cite{kamishima2003nantonac}, Jester~\cite{goldberg2001eigentaste}, and prioritized cancer gene expression data from The Cancer Genome Atlas (TCGA)~\cite{tcga}. The Sushi dataset consists of $5,000$ rankings of $N=10$ types of sushi. Each ranking has associated metadata indicating the region in Japan where the ranking originated. We convert the ordered lists into complete rankings (or permutations) and filter the data to contain rankings from regions with at least $20$ individuals to arrive at $10$ regions with a total of $M = 4,981$ samples. Each region is treated as a client (see the Supplement for details). 
Since no ground-truth information is available, we evaluate the performance of the methods using the Kemeny objective, normalized by the number of ranked entities. 
Borda still outperforms other methods (Figure~\ref{fig:real_data_sushi}). The Jester dataset consists of scores in the continuous interval $[-10,10]$ for $N=100$ jokes rated by $48,483$ individuals. We filter the dataset to include $14,116$ individuals who rated all $100$ jokes and converted the scores into complete rankings. The rankings are randomly split into $50$ groups corresponding to the clients (Figure~\ref{fig:real_data_jester}). The two prioritized (ranked) gene expression datasets LUAD and LUSC from TCGA consist of gene expression profiles of $501$ and $491$ patients, respectively. For prioritization, we select $N =978$ Landmark genes~\cite{subramanian2017next}. We randomly partition the resulting rankings among $10$ clients. The performance of our algorithms is shown in Figure~\ref{fig:real_data_cancer}. The results for the Maine Governor Election, involving $L=50$ clients, and $132,251$ rankings of $8$ candidates are in the Supplement. The code is available at \url{https://anonymous.4open.science/r/FRA-E1B2}. 

To implement FRA with quantization features for real data, we estimate $\phi$ via a simple search procedure and set it to $\phi=0.6$ (values in $[0.1, 0.8]$ offer similar performance). 


\bibliographystyle{plain}
\bibliography{FedRankAggre}
\clearpage
\newpage

\onecolumn
\section{\Large{Supplement}}
\section{Additional Proofs}
\subsection{Proof of Lemma \ref{lem:geometric}}
\begin{proof}
Fix $j$ and $i$ in $[N]$. 
We only consider the  case when $j<\sigma_0(i)$. For $j>\sigma_0(i)$, the proof follows similarly (and is obvious when $j=\sigma_0(i)$). 
Let $\mathcal{M}_{ij}=\{\sigma:\sigma\in \mathbb{S}_N,\sigma(i)=j\}$ be the set of permutations $\sigma$ such that $\sigma(i)=j$. We create an injective mapping $\mathcal{F}$ from $\mathcal{M}_{ij}$ to $\mathbb{S}_N$ such that $K_{\tau}(\sigma,\sigma_0)-K_{\tau}(\mathcal{F}(\sigma),\sigma_0)\ge \sigma_0(i)-j$ for any $\sigma\in \mathcal{M}_{ij}$. Then, $$Pr(\sigma\in\mathcal{M}_{ij})\le \phi^{\sigma_0(i)-j}Pr(\sigma\in Image(\mathcal{F}))\le \phi^{\sigma_0(i)-j},$$
where $Image(\mathcal{F})$ is the set of all output permutations of $\mathcal{F}$. 

For $\sigma\in\mathcal{M}_{ij}$, there exist at least $\sigma_0(i)-j$ indices $k$ such that $\sigma(\sigma^{-1}_0(k))>j=\sigma(i)$ and $k\in[\sigma_0(i)].$ 
This claim holds  because there are $\sigma_0(i)$ indices $k$ such that $k\in [\sigma_0(i)]$ while there are at most $j$ indices $k$ such that $\sigma(\sigma^{-1}_0(k))\le j$. Let $\sigma^{-1}_0(k_1),\ldots,\sigma^{-1}_0(k_{\sigma_0(i)-j}))$ be $\sigma_0(i)-j$ such indices with the minimum $\sigma$ values, i.e., we have $\sigma(\sigma^{-1}_0(k))>\sigma(\sigma^{-1}_0(k_\ell))$ for any $\ell\in[\sigma_0(i)-j]$, $k\in [\sigma_0(i)]$ and $\sigma(\sigma^{-1}_0(k))>j$. 
Note that $k_{\sigma_0(i)-j}<\sigma_0(i)$. Without loss of generality, assume that  $k_1<\ldots<k_{\sigma_0(i)-j}<k_{\sigma_0(i)-j+1}=\sigma_0(i)$.  
Define the function $\mathcal{F}(\sigma)=\sigma'\in\mathbb{S}_N$ as follows:
\begin{align}\label{eq:defsigmaprime}
    &\sigma'(k)\nonumber\\
    =&\begin{cases}
        \sigma(k),&\text{ if $k\ne \sigma^{-1}_0(k_\ell)$ for $\ell\in[\sigma_0(i)-j+1];$}\\
        \sigma(\sigma^{-1}_0(k_\ell)), &\text{ if $k=\sigma^{-1}_0(k_{\ell+1})$ for $\ell\in [\sigma_0(i)-j];$}\\
        \sigma(i),&\text{ if $k=\sigma^{-1}_0(k_1).$}
    \end{cases}
\end{align}
By definition,  $\mathcal{F}$ is an injective mapping. 
In what follows, we show that 
\begin{align}\label{eq:ktaudecrease}
K_\tau(\sigma,\sigma_0)-K_\tau(\sigma',\sigma_0)= \sigma_0(i)-j.    
\end{align}
Let 
\begin{align}
&I(j)=|\{\sigma^{-1}(j),\sigma^{-1}(t)\}:(\sigma(\sigma^{-1}(j))-\sigma(\sigma^{-1}(t)))\nonumber\\
&\cdot(\sigma_0(\sigma^{-1}(j))-\sigma_0(\sigma^{-1}(t)))<0,t\in[N]\backslash\{j\}\}|,\nonumber\\
&I'(j)=|\{(\sigma')^{-1}(j),(\sigma')^{-1}(t)\}:(\sigma'((\sigma')^{-1}(j))\nonumber\\
&-\sigma((\sigma')^{-1}(t)))\cdot(\sigma((\sigma')^{-1}(j))-\sigma((\sigma')^{-1}(t)))<0,\nonumber\\
&t\in[N]\backslash\{j\}\}|,\label{eq:ij}\\
&I(k)=|\{\sigma^{-1}(k),\sigma^{-1}(t)\}:(\sigma(\sigma^{-1}(k))-\sigma(\sigma^{-1}(t)))\nonumber\\
&\cdot(\sigma_0(\sigma^{-1}(k))-\sigma_0(\sigma^{-1}(t)))<0,t\in[N]\backslash\{j,k\}\}\nonumber,\\
&I'(k)=|\{(\sigma')^{-1}(k),(\sigma')^{-1}(t)\}:(\sigma'((\sigma')^{-1}(k))\nonumber\\
&-\sigma((\sigma')^{-1}(t)))(\sigma((\sigma')^{-1}(k))-\sigma((\sigma')^{-1}(t)))<0,\nonumber\\
&t\in[N]\backslash\{j\}\}|,\label{eq:ik}
\end{align}
for $k\in [N]\backslash\{j\}$. We prove \eqref{eq:ktaudecrease} by showing that $I(j)-I'(j)=\sigma_0(i)-j$ and $I(k)=I'(k)$ for $k\in [N]\backslash\{j\}$. By definition of $k_\ell$ above, 
$\ell\in [\sigma_0(i)-j]$, we have that $\sigma(\sigma^{-1}(j))=\sigma(i)<\sigma(\sigma^{-1}_0(k_\ell))$ and that  $\sigma_0(i)>\sigma_0(\sigma^{-1}_0(k_\ell))=k_\ell$ for $\ell\in [\sigma_0(i)-j]$. On the other hand, by the definition of $\sigma'$, we have that
$\sigma'((\sigma')^{-1}(j))<\sigma'(\sigma^{-1}_0(k_\ell))=\sigma(\sigma^{-1}_0(k_{\ell-1}))$ for $\ell\in\{2,\ldots,\sigma_0(i)-j+1\}$ and that $\sigma_0((\sigma')^{-1}(j))=\sigma_0(\sigma^{-1}_0(k_1))<\sigma_0(\sigma^{-1}_0(k_\ell)$ 
for $\ell\in\{2,\ldots,\sigma_0(i)-j+1\}$. In addition, for $t\notin\{\sigma(\sigma^{-1}_0(k_\ell))\}_{\ell\in [\sigma_0(i)-j+1]}$, we have that $t\notin\{\sigma'(\sigma^{-1}_0(k_\ell))\}_{\ell\in [\sigma_0(i)-j+1]}$ 
and that 
$\sigma^{-1}(t)=(\sigma')^{-1}(t)$ and hence, $\sigma_0(\sigma^{-1}(t))=\sigma_0((\sigma')^{-1}(t))$. Moreover, we have $k<j$ or $$k>\max_{\ell\in [\sigma_0(i)-j+1]}\sigma(\sigma^{-1}_0(k_\ell))=\max_{\ell\in [\sigma_0(i)-j+1]}\sigma'(\sigma^{-1}_0(k_\ell))$$ 
for any $k=\sigma(\sigma^{-1}_0(t))$, $t\in[\sigma_0(i)]$. Therefore, 
we have
\begin{align*}
    &|\{t:(\sigma(\sigma^{-1}(j))-\sigma(\sigma^{-1}(t)))\cdot(\sigma_0(\sigma^{-1}(j))\nonumber\\
    &-\sigma_0(\sigma^{-1}(t)))<0,t\notin\{\sigma(\sigma^{-1}_0(k_\ell))\}_{\ell\in [\sigma_0(i)-j+1]}\}|\\
    =&|\{t:(\sigma'((\sigma')^{-1}(j))-\sigma'((\sigma')^{-1}(t)))\cdot(\sigma_0(\sigma^{-1}(j))\nonumber\\
    &-\sigma_0(\sigma^{-1}(t)))<0,t\notin\{\sigma'(\sigma^{-1}_0(k_\ell))\}_{\ell\in [\sigma_0(i)-j+1]}\}|.
\end{align*}
This proves \eqref{eq:ij}. 
In addition, 
\begin{align}\label{eq:k}
&|\{t:(\sigma'((\sigma')^{-1}(k))-\sigma'((\sigma')^{-1}(t)))\cdot(\sigma_0(\sigma^{-1}(k))\nonumber\\
&-\sigma_0(\sigma^{-1}(t)))<0,t\in\{\sigma(\sigma^{-1}_0(k_\ell))\}_{\ell\in [\sigma_0(i)-j]}\}| \\
=&|\{t:(\sigma((\sigma)^{-1}(k))-\sigma((\sigma)^{-1}(t)))\cdot(\sigma_0(\sigma^{-1}(k))\nonumber\\
&-\sigma_0(\sigma^{-1}(t)))<0,t\in\{\sigma(\sigma^{-1}_0(k_\ell))\}_{\ell\in [\sigma_0(i)-j]}\}|,
\end{align}
for $k\notin \{\sigma(\sigma^{-1}_0(k_\ell))\}_{\ell\in [\sigma_0(i)-j+1]}\}$ and $\sigma_0(\sigma^{-1}(k))\le \sigma_0(i)$. Similarly, we have \eqref{eq:k} for $\sigma_0(\sigma^{-1}(k))> \sigma_0(i)$. By the definition of $\sigma'$, we have 
\begin{align}\label{eq:k2}
&|\{t:(\sigma'((\sigma')^{-1}(k))-\sigma'((\sigma')^{-1}(t)))\cdot(\sigma_0(\sigma^{-1}(k))\nonumber\\
&-\sigma_0(\sigma^{-1}(t)))<0,t\notin\{\sigma(\sigma^{-1}_0(k_\ell))\}_{\ell\in [\sigma_0(i)-j]}\}|\\ 
=&|\{t:(\sigma((\sigma)^{-1}(k))-\sigma((\sigma)^{-1}(t)))\cdot(\sigma_0(\sigma^{-1}(k))\nonumber\\
&-\sigma_0(\sigma^{-1}(t)))<0,t\notin\{\sigma(\sigma^{-1}_0(k_\ell))\}_{\ell\in [\sigma_0(i)-j]}\}|,
\end{align}
for $k\notin\{\sigma(\sigma^{-1}_0(k_\ell))\}_{\ell\in [\sigma_0(i)-j]},$ and 
\begin{align}\label{eq:k3}
&|\{t:(\sigma'((\sigma')^{-1}(k))-\sigma'((\sigma')^{-1}(t)))\cdot(\sigma_0(\sigma^{-1}(k))\nonumber\\
&-\sigma_0(\sigma^{-1}(t)))<0,t\in\{\sigma(\sigma^{-1}_0(k_\ell))\}_{\ell\in [\sigma_0(i)-j]}\}|\\ 
=&|\{t:(\sigma((\sigma)^{-1}(k))-\sigma((\sigma)^{-1}(t)))\cdot(\sigma_0(\sigma^{-1}(k))\nonumber\\
&-\sigma_0(\sigma^{-1}(t)))<0,t\in\{\sigma(\sigma^{-1}_0(k_\ell))\}_{\ell\in [\sigma_0(i)-j]}\}|,
\end{align}
for $k\in \{\sigma(\sigma^{-1}_0(k_\ell))\}_{\ell\in [\sigma_0(i)-j]}\}$. Equations~\eqref{eq:k},~\eqref{eq:k2}, and~\eqref{eq:k3} imply~\eqref{eq:ik}. Therefore, we have
\begin{align*}
    K_{\tau}(\sigma,\sigma_0)=&I(j)+\frac{\sum_{k\in[N]\backslash\{j\}}I(k)}{2}\\
    =&I'(j)+\frac{\sum_{k\in[N]\backslash\{j\}}I'(k)}{2}+\sigma_0(i)-j.\\
    =&K_\tau(\sigma',\sigma_0)+\sigma_0(i)-j.
\end{align*}
\end{proof}
\subsection{Proof of Lemma \ref{lem:exporder}}
\begin{proof}
It suffices to show that $E[\sigma(\sigma^{-1}_0(i+1))]>E[\sigma(\sigma^{-1}_0(i))]$ for any $i\in[N-1]$. 
Let $T_{ij}=1$ if $\sigma(i)>\sigma(j)$ and $T_{ij}=0$ otherwise, for $i,j\in[N]$. Then, $\sigma(i)=1+\sum^N_{j=1}T_{ij}$ for $i\in [N]$. In what follows, we show that 
\begin{align}\label{eq:expequal}
E[T_{\sigma^{-1}_0(i_1)\sigma^{-1}_0(j_1)}]=E[T_{\sigma^{-1}_0(i_2)\sigma^{-1}_0(j_2)}]=E[T^{i_1-j_1}_{(i_1-j_1+1)1}]
\end{align} 
for any $i_1>j_1$, $i_2>j_2$, and $i_1-j_1=i_2-j_2$, where $T^{i_1-j_1}_{(i_1-j_1+1)1}=1$ if $\sigma'(i_1-j_1+1)>\sigma'(1)$ for a permutation $\sigma'\in\mathbb{S}_{i_1-j_1+1}$ randomly sampled from the Mallows distribution with centroid permutation being the identity permutation with length $i_1-j_1+1$.
\begin{align*}
&|\{\{\sigma^{-1}_0(k),\sigma^{-1}_0(k')\}:
(\sigma(\sigma^{-1}_0(k))-\sigma(\sigma^{-1}_0(k')))\nonumber\\
&\cdot(\sigma_0(\sigma^{-1}_0(k))-\sigma_0(\sigma^{-1}_0(k')))<0, 
k'\in\{j_1,\ldots,i_1\}\}|    
\end{align*}
is constant when fixing the values $\sigma(k)$ for $k=\sigma^{-1}_0(\ell)$,  $\ell\in[N]\backslash\{j_1,\ldots,i_1\}$. Therefore, the probability that $\sigma(\sigma^{-1}_0(i_1))>\sigma(\sigma^{-1}_0(j_1))$ conditioned on the values $\sigma(k)$ for $k=\sigma^{-1}_0(\ell)$ $\ell\in[N]\backslash\{j_1,\ldots,i_1\}$ is the same as $\sigma^{\{\sigma^{-1}_0(i)\}_{i\in\{j_1,\ldots,i_1\}}}(1+i_1-j_1)>\sigma^{\{\sigma^{-1}_0(i)\}_{i\in\{j_1,\ldots,i_1\}}}(1)$, which is $E[T^{i_1-j_1}_{(i_1-j_1+1)1}]$. Taking expectations over all choices of $\sigma(k)$ for, $k=\sigma^{-1}_0(\ell)$ $\ell\in[N]\backslash\{j_1,\ldots,i_1\}$, we have \eqref{eq:expequal}. Then, we have that
\begin{align}\label{eq:i+1iexp}
    &E[\sigma(\sigma^{-1}_0(i+1))]-E[\sigma(\sigma^{-1}_0(i))]\nonumber\\
=&E[T_{\sigma^{-1}_0(i+1)\sigma^{-1}_0(1)}]+\sum^{N}_{j=2}E[T_{\sigma^{-1}_0(i+1)\sigma^{-1}_0(j)}]\nonumber\\
&-\sum^{N-1}_{j=1}E[T_{\sigma^{-1}_0(i)\sigma^{-1}_0(j)}]-E[T_{\sigma^{-1}_0(i)\sigma^{-1}_0(N)}]\nonumber\\
=&E[T_{\sigma^{-1}_0(i+1)\sigma^{-1}_0(1)}]-E[T_{\sigma^{-1}_0(i)\sigma^{-1}_0(N)}].
\end{align}
Note that swapping $\sigma(\sigma^{-1}_0(i_1))$ and $\sigma(\sigma^{-1}_0(i_2))$ reduces the  Kendall-Tau distance from $\sigma_0$ by at least $1$ for $i_1>i_2$ and $\sigma(\sigma^{-1}_0(i_1))<\sigma(\sigma^{-1}_0(i_2))$, i.e., 
$K_\tau(\sigma,\sigma_0)-K_\tau(\sigma',\sigma_0)\ge 1$, where $\sigma'(\sigma^{-1}_0(i_1))=\sigma(\sigma^{-1}_0(i_2))$, $\sigma'(\sigma^{-1}_0(i_2))=\sigma(\sigma^{-1}_0(i_1))$, and $\sigma'(k)=\sigma(k)$ for $k\in[N]\backslash\{\sigma^{-1}_0(i_1),\sigma^{-1}_0(i_2)\}$. Hence, we have that $E[T_{\sigma^{-1}_0(i+1)\sigma^{-1}_0(1)}]\ge \frac{1}{1+\phi}$ and that $E[T_{\sigma^{-1}_0(i)\sigma^{-1}_0(N)}]\le \frac{\phi}{1+\phi}$. Therefore, by~\eqref{eq:i+1iexp},
\begin{align}\label{eq:i+1greaterthani}
E[\sigma(\sigma^{-1}_0(i+1))]-E[\sigma(\sigma^{-1}_0(i))]\ge \frac{1-\phi}{1+\phi},     
\end{align}
which completes the proof.
\end{proof}
\begin{remark}
We describe next how to compute $E[\sigma(\sigma^{-1}_0(i))]$ recursively. To this end, we need to compute $E[T^{i}_{(i+1)i}]$ for each $i\in[N-1]$. When $i=1$, we have $E[T^{1}_{21}]=\frac{\phi}{1+\phi}$. For $i>1$, we have
\begin{align}\label{eq:computeei1}
 E[T^{i}_{(i+1)1}]=\frac{\phi^{i}+\sum^{i-1}_{j=1}\phi^j E[T^{i-1}_{i1}]}{\sum^{i}_{j=1}\phi^j}  
\end{align}
Then, one can compute $E[T_{\sigma^{-1}_0(i+1)\sigma^{-1}_0(1)}]=E[T^{i}_{(i+1)1}]$ and $E[T_{\sigma^{-1}_0(i)\sigma^{-1}_0(N)}]=E[T^{N-i}_{(N-i+1)1}]$ recursively based on \eqref{eq:computeei1} and thus 
\begin{align}\label{eq:computeej}
E[\sigma(\sigma^{-1}_0(i))]=\sum^i_{j=1} (E[T_{\sigma^{-1}_0(j+1)\sigma^{-1}_0(1)}]-E[T_{\sigma^{-1}_0(j)\sigma^{-1}_0(N)}])   
\end{align}
based on \eqref{eq:computeei1}.
\end{remark}
\subsection{Proof of Theorem \ref{thm:averaging}}
\begin{proof}
We first show that 
\begin{align}\label{eq:concentrationbound}
&Pr(\sum^M_{m=1}\sigma_m(\sigma^{-1}_0(i)))-\sum^M_{m=1}E[\sigma_m(\sigma^{-1}_0(i)))]>rM)\nonumber\\
\le& e^{-\frac{r^2M}{4C^2}}
\end{align}
for any $r>0$. 
According to the Chernoff bound,
\begin{align}\label{eq:chernoff}
&Pr(\sum^M_{m=1}\sigma_m(\sigma^{-1}_0(i)))-\sum^M_{m=1}E[\sigma_m(\sigma^{-1}_0(i)))]>rM)\nonumber\\
\le& E[e^{\lambda \big (\sum^M_{m=1}\sigma_m(\sigma^{-1}_0(i)))-\sum^M_{m=1}E[\sigma_m(\sigma^{-1}_0(i)))]\big)}]e^{-\lambda rM}\nonumber\\
=&\big(E[e^{\lambda(\sigma_1(\sigma^{-1}_0(i)))-E[\sigma_1(\sigma^{-1}_0(i)))])}]\big)^Me^{-\lambda rM}\nonumber\\
\overset{(a)}{\le} &\big(E[e^{\lambda(\sigma_1(\sigma^{-1}_0(i))-\sigma_2(\sigma^{-1}_0(i)))}]\big)^Me^{-\lambda rM}\nonumber\\
=&(\sum_{|j|\ge C,j\in\{-N,\ldots,N\}}Pr(\sigma_1(\sigma^{-1}_0(i))-\sigma_2(\sigma^{-1}_0(i))=j)e^{\lambda j}\nonumber\\
&+\sum_{|j|< C,j\in\{-N,\ldots,N\}}\nonumber\\
&Pr(\sigma_1(\sigma^{-1}_0(i))-\sigma_2(\sigma^{-1}_0(i))=j)e^{\lambda j})^Me^{-\lambda rM}\nonumber\\
\overset{(b)}{\le} &(\frac{8\phi^{\frac{C}{2}}e^{\lambda C}}{1-e^{\lambda}\phi^{\frac{1}{2}}}+\sum_{|j|< C,j\in\{-N,\ldots,N\}}\nonumber\\
&Pr(\sigma_1(\sigma^{-1}_0(i))-\sigma_2(\sigma^{-1}_0(i))=j)e^{\lambda j})^Me^{-\lambda rM}\nonumber\\
\overset{(c)}{\le} &(\frac{8\phi^{\frac{C}{2}}e^{\lambda C}}{1-e^{\lambda}\phi^{\frac{1}{2}}}+e^{\frac{C^2\lambda^2}{2}})^Me^{-\lambda rM}
\end{align}
for any positive $\lambda$, where $(a)$ follows from the convexity of the function $f(x)=e^{-\lambda x}$, $(b)$ follows from Lemma ~\ref{lem:geometric} and the fact that either $|\sigma_1(\sigma^{-1}_0(i))-i|\ge \frac{j}{2}$ or $|\sigma_2(\sigma^{-1}_0(i))-i|\ge \frac{j}{2}$ for $\sigma_1(\sigma^{-1}_0(i))-\sigma_2(\sigma^{-1}_0(i))=j$, and $(c)$ follows from the 
the fact that $Pr(\sigma_1(\sigma^{-1}_0(i))-\sigma_2(\sigma^{-1}_0(i))=j)=Pr(\sigma_1(\sigma^{-1}_0(i))-\sigma_2(\sigma^{-1}_0(i))=-j)$ and the fact that $\frac{(e^{-x}+e^x)}{2}\le e^{\frac{x^2}{2}}$ for any $x$. We choose 
\begin{align}\label{eq:Cr}
C= \max\{\frac{8\ln \frac{32}{r^2(1-\sqrt{\phi})}}{\ln \frac{1}{\phi}}, e^{\frac{\ln \frac{1}{\phi}}{16}},\sqrt{\frac{4r}{\ln \frac{1}{\phi}}}\}
\end{align}
In addition, we choose $\lambda$ such that 
\begin{align}\label{eq:lambda}
\lambda = \frac{r}{C^2}.  
\end{align}
From \eqref{eq:lambda}, we then have 
\begin{align*}
e^{\frac{C^2\lambda^2}{2}}e^{-\lambda r}  = e^{-\frac{r^2}{2C^2}}.
\end{align*}
In what follows, we show that 
\begin{align}\label{eq:firstterm}
\frac{8\phi^{\frac{C}{2}}e^{\lambda C}}{1-e^{\lambda}\phi^{\frac{1}{2}}}e^{-\lambda r} \le e^{-\frac{r^2}{4C^2}}-e^{-\frac{r^2}{2C^2}},  
\end{align}
which is equivalent to
\begin{align*}
    \frac{8\phi^{\frac{C}{2}}e^{\lambda C}}{1-e^{\lambda}\phi^{\frac{1}{2}}}\le e^{\frac{3r^2}{4C^2}}-e^{\frac{r^2}{2C^2}}.
\end{align*}
Note that $e^{a}-e^{b}\ge a-b$ for any $a>b>0$. It suffices to show that
\begin{align}\label{eq1}
    \frac{8\phi^{\frac{C}{2}}e^{\lambda C}}{1-e^{\lambda}\phi^{\frac{1}{2}}}\le \frac{r^2}{4C^2}.
\end{align}
Since from \eqref{eq:Cr} and \eqref{eq:lambda}, we have
\begin{align*}
\lambda\le \frac{\ln\frac{1}{\phi}}{4},
\end{align*}
which implies that 
\begin{align*}
\frac{8\phi^{\frac{C}{2}}e^{\lambda C}}{1-e^{\lambda}\phi^{\frac{1}{2}}}\le \frac{8\phi^{\frac{C}{4}}}{1-\sqrt{\phi}}.    
\end{align*}
Next, we show that 
\begin{align}\label{eq2}
   \frac{8\phi^{\frac{C}{4}}}{1-\sqrt{\phi}}\le \frac{r^2}{4C^2},
\end{align}
which implies \eqref{eq1} and thus \eqref{eq:firstterm}. Taking logarithms of both sides, \eqref{eq2} is equivalent to 
\begin{align}\label{eq3}
    2\ln C +\ln \frac{32}{r^2(1-\sqrt{\phi})}\le \frac{C\ln \frac{1}{\phi}}{4}.
\end{align}
From \eqref{eq:C}, we have $$\frac{C\ln \frac{1}{\phi}}{8}\ge \ln \frac{32}{r^2(1-\sqrt{\phi})}$$ and $$ \frac{C\ln \frac{1}{\phi}}{8}\ge 2\ln C,$$
the sum of which implies \eqref{eq3}. Hence we have \eqref{eq:concentrationbound}. Similarly, we have 
\begin{align}\label{eq:concentrationbound1}
&Pr(\sum^M_{m=1}\sigma_m(\sigma^{-1}_0(i)))-\sum^M_{m=1}E[\sigma_m(\sigma^{-1}_0(i)))]<-rM)\nonumber\\
\le& e^{-\frac{r^2M}{4C^2}}.
\end{align}
Hence, 
\begin{align*}
&Pr(|\sum^M_{m=1}\sigma_m(\sigma^{-1}_0(i)))-\sum^M_{m=1}E[\sigma_m(\sigma^{-1}_0(i)))]|>rM)\\
\le& 2e^{-\frac{r^2M}{4C^2}}.    
\end{align*}
According to \eqref{eq:i+1greaterthani}, we choose  
$r = \frac{(1-\phi)}{2(1+\phi)}$. Hence the probability that there exists some $i\in[N]$ such that 
\begin{align*}
&Pr(|\frac{\sum^M_{m=1}\sigma_m(\sigma^{-1}_0(i)))-\sum^M_{m=1}E[\sigma_m(\sigma^{-1}_0(i)))]}{M}|\\
>&\max\{\frac{E[\sigma_m(\sigma^{-1}_0(i)))-E[\sigma_m(\sigma^{-1}_0(i-1)))}{2},\\
&\frac{E[\sigma_m(\sigma^{-1}_0(i+1)))-E[\sigma_m(\sigma^{-1}_0(i)))}{2}\})    
\end{align*}
is at most $\delta$ whenever~\eqref{eq:lowerboundM} holds. Thus, $\hat{\sigma}_0$ equals $\sigma_0$ with probability at least $1-\delta$.
\end{proof}
\subsection{Proof of Theorem \ref{thm:fedaverage}}
\begin{proof}
We first show that $E[\hat{\sigma}_\ell(\sigma^{-1}_0(i+1))]-E[\hat{\sigma}_\ell(\sigma^{-1}_0(i))]\ge \frac{1}{2}$. Note that 
\begin{align}\label{eq:expofAell}
&E[\hat{\sigma}_\ell(\sigma^{-1}_0(i))]\nonumber\\
=&\sum^{N}_{j=1}Pr(A_\ell(\sigma^{-1}_0(i))\ge \frac{E[\sigma(\sigma^{-1}_0(j))+\sigma(\sigma^{-1}_0(j-1))]}{2}) 
\end{align}
for any $i\in [N]$, 
where $\sigma$ follows the distribution \eqref{eq:mallow} and $\sigma(\sigma^{-1}_0(0))=-N$ is assumed to satisfy  $A_\ell(\sigma^{-1}_0(i))\ge\frac{\sigma(\sigma^{-1}_0(1))+\sigma(\sigma^{-1}_0(0))}{2}$ for $i\in [N]$. Next, we show that 
\begin{align}\label{eq:accumulate}
&Pr(A_\ell(\sigma^{-1}_0(i+1))\ge \frac{E[\sigma(\sigma^{-1}_0(j))+\sigma(\sigma^{-1}_0(j-1))]}{2})\nonumber\\
\ge &Pr(A_\ell(\sigma^{-1}_0(i))\ge \frac{E[\sigma(\sigma^{-1}_0(j))+\sigma(\sigma^{-1}_0(j-1))]}{2})    
\end{align}
for $i\in[N-1]$, which is equivalent to
\begin{align}\label{eq:accumulate1}
  Pr(\sum^{m_\ell}_{m=1}&\sigma_{\ell,m}(\sigma^{-1}_0(i+1))\nonumber\\
  &\ge \frac{m_\ell E[\sigma(\sigma^{-1}_0(j))+\sigma(\sigma^{-1}_0(j-1))]}{2} )\nonumber\\ 
  \ge   Pr(\sum^{m_\ell}_{m=1}&\sigma_{\ell,m}(\sigma^{-1}_0(i))\nonumber\\
  &\ge \frac{m_\ell E[\sigma(\sigma^{-1}_0(j))+\sigma(\sigma^{-1}_0(j-1))]}{2} ). 
\end{align}
We prove by induction on $m_\ell$ that 
\begin{align}\label{eq:accumulate2}
Pr(\sum^{m_\ell}_{m=1}&\sigma_{\ell,m}(\sigma^{-1}_0(i+1))\nonumber\\
&\ge x |\sigma_{\ell,m}(\sigma^{-1}_0(j)),j\in [N]\backslash\{i,i+1\})\nonumber\\ 
  \ge   Pr(\sum^{m_\ell}_{m=1}&\sigma_{\ell,m}(\sigma^{-1}_0(i))\nonumber\\
  &\ge x |\sigma_{\ell,m}(\sigma^{-1}_0(j)),j\in [N]\backslash\{i,i+1\}), 
\end{align}
for any integers $\ell\in [L]$ and   $m\in[m_\ell]$ and  real value  $x\in\mathbb{R}$. In the proof of~\eqref{eq:accumulate2}, we assume that all the probabilities are conditioned on the values of $\sigma_{\ell,m}(\sigma^{-1}_0(j)),j\in [N]\backslash\{i,i+1\}$ and omit the conditioning for ease of notation. 
For $m_\ell=1$, there are two choices for  $\sigma_{\ell,1}$ when $\sigma_{\ell,1}(\sigma^{-1}_0(j))$, $j\in [N]\backslash\{i,i+1\}$ are fixed. Note that $Pr(\sigma_{\ell,1}(\sigma^{-1}_0(i))>\sigma_{\ell,1}(\sigma^{-1}_0(i+1)))=\phi Pr(\sigma_{\ell,1}(\sigma^{-1}_0(i+1))>\sigma_{\ell,1}(\sigma^{-1}_0(i+1)))$. Hence, \eqref{eq:accumulate2} holds. Suppose \eqref{eq:accumulate2} holds for $m_\ell=m'$. For $m_\ell=m'+1$, there are two choices $( \sigma_{\ell,m'+1}(\sigma^{-1}_0(i+1))),\sigma_{\ell,m'+1}(\sigma^{-1}_0(i))) )= (a,b)$ and $( \sigma_{\ell,m'+1}(\sigma^{-1}_0(i+1))),\sigma_{\ell,m'+1}(\sigma^{-1}_0(i))) )= (b,a)$ for  $\sigma_{\ell,m'+1},$ where $a,b\in[N]$ and $a>b$. We then have 
\begin{align}\label{eq:accumulate3}
&Pr(\sum^{m_\ell}_{m=1}\sigma_{\ell,m}(\sigma^{-1}_0(i+1))\ge x )\nonumber\\ 
  =   &Pr(\sigma_{\ell,m'+1}(\sigma^{-1}_0(i+1)=a)\nonumber\\
  &\cdot Pr(\sum^{m'}_{m=1}\sigma_{\ell,m}(\sigma^{-1}_0(i+1))\ge x-a)\nonumber\\
  &+Pr(\sigma_{\ell,m'+1}(\sigma^{-1}_0(i+1)=b)\nonumber\\
  &\cdot Pr(\sum^{m'}_{m=1}\sigma_{\ell,m}(\sigma^{-1}_0(i+1))\ge x-b)\\
  =&Pr(\sigma_{\ell,m'+1}(\sigma^{-1}_0(i+1)=a)\nonumber\\
  &\cdot Pr(\sum^{m'}_{m=1}\sigma_{\ell,m}(\sigma^{-1}_0(i+1))\ge x-a)\nonumber\\
  &+\phi Pr(\sigma_{\ell,m'+1}(\sigma^{-1}_0(i+1)=a)\nonumber\\
  &\cdot Pr(\sum^{m'}_{m=1}\sigma_{\ell,m}(\sigma^{-1}_0(i+1))\ge x-b)
\end{align}
Similarly, we have 
\begin{align*}
&Pr(\sum^{m_\ell}_{m=1}\sigma_{\ell,m}(\sigma^{-1}_0(i))\ge x )\\ 
  =&\phi Pr(\sigma_{\ell,m'+1}(\sigma^{-1}_0(i)=b)\\
  &\cdot Pr(\sum^{m'}_{m=1}\sigma_{\ell,m}(\sigma^{-1}_0(i))\ge x-a)\nonumber\\
  &+Pr(\sigma_{\ell,m'+1}(\sigma^{-1}_0(i)=b)\\
  &\cdot Pr(\sum^{m'}_{m=1}\sigma_{\ell,m}(\sigma^{-1}_0(i))\ge x-b). 
\end{align*}
Note that $Pr(\sigma_{\ell,m'+1}(\sigma^{-1}_0(i)=b)=Pr(\sigma_{\ell,m'+1}(\sigma^{-1}_0(i+1)=a)$ and by the induction hypothesis that
\begin{align*}
&Pr(\sum^{m'}_{m=1}\sigma_{\ell,m}(\sigma^{-1}_0(i))\ge x-b)\\
\le &Pr(\sum^{m'}_{m=1}\sigma_{\ell,m}(\sigma^{-1}_0(i+1))\ge x-b),\mbox{ and}\\
&Pr(\sum^{m'}_{m=1}\sigma_{\ell,m}(\sigma^{-1}_0(i))\ge x-a)\\
\le &Pr(\sum^{m'}_{m=1}\sigma_{\ell,m}(\sigma^{-1}_0(i+1))\ge x-a),
\end{align*}
from which we have
\begin{align*}
 & Pr(\sum^{m_\ell}_{m=1}\sigma_{\ell,m}(\sigma^{-1}_0(i+1))\ge x )\\
 &-Pr(\sum^{m_\ell}_{m=1}\sigma_{\ell,m}(\sigma^{-1}_0(i))\ge x ) \\
 \ge&(1-\phi) Pr(\sigma_{\ell,m'+1}(\sigma^{-1}_0(i)=a) \\
 &\cdot Pr(\sum^{m'}_{m=1}\sigma_{\ell,m}(\sigma^{-1}_0(i+1))\ge x-a)\\
 &-Pr(\sum^{m'}_{m=1}\sigma_{\ell,m}(\sigma^{-1}_0(i))\ge x-b))\\
 \ge& 0.
\end{align*}
Hence, we have that ~\eqref{eq:accumulate2} holds for $m_\ell=m'+1$ and therefore for all positive integers $m_\ell$ by induction. Thus, the result~\eqref{eq:accumulate} follows. Combining the above result with ~\eqref{eq:expofAell}, we have that
\begin{align}\label{eq:expdiff}
&E[A_\ell(\sigma^{-1}_0(i+1))]-
E[A_\ell(\sigma^{-1}_0(i))]\nonumber\\
\ge 
&Pr(A_\ell(\sigma^{-1}_0(i+1))\ge \frac{E[\sigma(\sigma^{-1}_0(i+1))+\sigma(\sigma^{-1}_0(i))]}{2})\nonumber\\
&- Pr(A_\ell(\sigma^{-1}_0(i))\ge \frac{E[\sigma(\sigma^{-1}_0(i+1))+\sigma(\sigma^{-1}_0(i))]}{2}).
\end{align}
According to~\eqref{eq:i+1greaterthani} and~\eqref{eq:concentrationbound}, we also have 
\begin{align}\label{eq:iupperbound}
&Pr(A_\ell(\sigma^{-1}_0(i))\ge \frac{E[\sigma(\sigma^{-1}_0(i+1))+\sigma(\sigma^{-1}_0(i))]}{2})\nonumber\\
=&Pr\Big(  \sum^{m_\ell}_{m=1}\sigma_{\ell,m}(\sigma^{-1}_0(i))-  \sum^{m_\ell}_{m=1}E[\sigma_{\ell,m}(\sigma^{-1}_0(i))]\nonumber\\
&>  \frac{m_\ell E[\sigma(\sigma^{-1}_0(i+1))-\sigma(\sigma^{-1}_0(i))]}{2}\Big)\nonumber\\ 
\le & e^{-\frac{m_\ell(1-\phi)^2}{16C^2(1+\phi)^2}}.
\end{align}
According to~\eqref{eq:concentrationbound1}, it also holds that
\begin{align}\label{eq:i+1lowerbound}
&Pr(A_\ell(\sigma^{-1}_0(i+1))\ge \frac{E[\sigma(\sigma^{-1}_0(i+1))+\sigma(\sigma^{-1}_0(i))]}{2})\nonumber\\
=&1-Pr\Big(  \sum^{m_\ell}_{m=1}\sigma_{\ell,m}(\sigma^{-1}_0(i+1))\nonumber\\
&-  \sum^{m_\ell}_{m=1}E[\sigma_{\ell,m}(\sigma^{-1}_0(i+1))]\nonumber\\
&<  \frac{-m_\ell E[\sigma(\sigma^{-1}_0(i+1))-\sigma(\sigma^{-1}_0(i))]}{2}\Big)\nonumber\\ 
\le &1- e^{-\frac{m_\ell(1-\phi)^2}{16C^2(1+\phi)^2}}.
\end{align}
Note that $m_\ell\ge \frac{128C^2(1+\phi)^2\ln 4}{(1-\phi)^2}$. Then,~\eqref{eq:iupperbound} and~\eqref{eq:i+1lowerbound} imply 
\begin{align*}
&Pr\Big(A_\ell(\sigma^{-1}_0(i+1))
\ge \frac{E[\sigma(\sigma^{-1}_0(i+1))+\sigma(\sigma^{-1}_0(i))]}{2}\Big)\\
&-Pr\Big(A_\ell(\sigma^{-1}_0(i))\ge \frac{E[\sigma(\sigma^{-1}_0(i+1))+\sigma(\sigma^{-1}_0(i))]}{2}\Big)\\
\ge &\frac{127}{128}.    
\end{align*}
From~\eqref{eq:expdiff}, we have $E[\hat{\sigma}_\ell(\sigma^{-1}_0(i+1))]-
E[\hat{\sigma}_\ell(\sigma^{-1}_0(i))]\ge \frac{127}{128}$. Moreover, for two independent and identically distributed random variables $\hat{\sigma}_\ell(\sigma^{-1}_0(i)$ and $\hat{\sigma'}_\ell(\sigma^{-1}_0(i)$, 
we have
\begin{align}\label{eq:devfromexp}  &Pr(\hat{\sigma}_\ell(\sigma^{-1}_0(i))-\hat{\sigma'}_\ell(\sigma^{-1}_0(i))\ge j)\nonumber\\
\overset{(a)}{\le}&
2(Pr\Big(A_\ell(\sigma^{-1}_0(i))-E[A_\ell(\sigma^{-1}_0(i))]\ge \frac{(2j-1)(1-\phi)}{4(1+\phi)}\Big)\nonumber\\
&+Pr\Big(A_\ell(\sigma^{-1}_0(i))-E[A_\ell(\sigma^{-1}_0(i))]\le \nonumber\\
&-\frac{(2j-1)(1-\phi)}{4(1+\phi)})\Big)\nonumber\\
\le& 
4e^{-\frac{m_\ell(\frac{2j-1}{4})^2(1-\phi)^2}{4(1+\phi)^2C^2}},
\end{align}
where $(a)$ uses the fact~\eqref{eq:i+1greaterthani}. According to the Chernoff bound,
\begin{align}\label{eq:chernoff}
&Pr(\sum^L_{\ell=1}\hat{\sigma}_\ell(\sigma^{-1}_0(i)))-\sum^L_{\ell=1}E[\hat{\sigma}_\ell(\sigma^{-1}_0(i)))]>\frac{127L}{256})\nonumber\\
\le& E[e^{\lambda \big (\sum^L_{\ell=1}\hat{\sigma}_\ell(\sigma^{-1}_0(i)))-\sum^L_{\ell=1}E[\hat{\sigma}_\ell(\sigma^{-1}_0(i)))]\big)}]e^{-\lambda \frac{127L}{256}}\nonumber\\
=&\prod^L_{\ell=1}\big(E[e^{\lambda(\hat{\sigma}_\ell(\sigma^{-1}_0(i)))-E[\hat{\sigma}_\ell(\sigma^{-1}_0(i)))])}]e^{- \frac{127\lambda}{256}}\big)\nonumber\\
\le &\prod^L_{\ell=1}\big(E[e^{\lambda(\hat{\sigma}_\ell(\sigma^{-1}_0(i)))-\hat{\sigma'}_\ell(\sigma^{-1}_0(i)))])}]e^{- \frac{127\lambda}{256}}\big)\nonumber\\
=&\prod^L_{\ell=1}((\sum_{|j|\ge 1,j\in\{-N,\ldots,N\}}\nonumber\\
&Pr(\hat{\sigma}_\ell(\sigma^{-1}_0(i))-\hat{\sigma'}_\ell(\sigma^{-1}_0(i))=j)e^{\lambda j}\nonumber\\
&+Pr(\hat{\sigma}_\ell(\sigma^{-1}_0(i))-\hat{\sigma'}_\ell(\sigma^{-1}_0(i))=0))e^{- \frac{127\lambda}{256}})\nonumber\\
\overset{(a)}{\le} &
e^{ \frac{-127\lambda L}{256}}\prod^L_{\ell = 1}(\frac{1}{1-4e^{\lambda-\frac{m_\ell(\frac{2j-1}{4})^2(1-\phi)^2}{4(1+\phi)^2C^2}}}),
\end{align}
where $(a)$ follows from \eqref{eq:devfromexp}. 
Choosing $\lambda= \frac{m_\ell(1-\phi)^2}{128C^2(1+\phi)^2}$, we have that
\begin{align*}
&Pr(\sum^L_{\ell=1}\hat{\sigma}_\ell(\sigma^{-1}_0(i)))-\sum^L_{\ell=1}E[\hat{\sigma}_\ell(\sigma^{-1}_0(i)))]>\frac{127L}{256})\\
\le &e^{ \frac{m_\ell(1-\phi)^2L}{260C^2(1+\phi)^2}}(\frac{4}{3})^L,
\end{align*}
whenever $m_\ell\ge \frac{256(1+\phi)^2}{(1-\phi)^2}$. Furthermore, when $m_\ell\ge \frac{C^2(260\ln(\frac{2N}{\delta}+\ln(\frac{4}{3}L))(1+\phi)^2}{L(1-\phi)^2}$, we have that $Pr(\sum^L_{\ell=1}\hat{\sigma}_\ell(\sigma^{-1}_0(i)))-\sum^L_{\ell=1}E[\hat{\sigma}_\ell(\sigma^{-1}_0(i)))]>\frac{127L}{256})
\le \frac{\delta}{2N}$. Similarly, we have $Pr(\sum^L_{\ell=1}\hat{\sigma}_\ell(\sigma^{-1}_0(i)))-\sum^L_{\ell=1}E[\hat{\sigma}_\ell(\sigma^{-1}_0(i)))]<\frac{-127L}{256})
\le \frac{\delta}{2N}$ when $m_\ell \ge \max\{\frac{256(1+\phi)^2}{(1-\phi)^2},\frac{C^2(260\ln(\frac{2N}{\delta}+\ln(\frac{4}{3}L))(1+\phi)^2}{L(1-\phi)^2}\}$. Hence, the probability that there exists some $i\in[N]$ satisfying $|\sum^L_{\ell=1}\hat{\sigma}_\ell(\sigma^{-1}_0(i)))-\sum^L_{\ell=1}E[\hat{\sigma}_\ell(\sigma^{-1}_0(i)))]|\ge \frac{-127L}{256}$ is at most $\delta$. Therefore, the server recovers $\sigma_0$ with probability at least $1-\delta$.
\end{proof}
\subsection{Proof of Lemma \ref{lem:majoritylehmer}}
\begin{proof} 
We first present a result from
~\cite{li2017efficient} showing that the probability of the coordinates of Lehmer codes decays exponentially with the deviation from the true values. Before that, we define the notation $\sigma^A\in \mathbb{S}_{|A|}$ for a permutation $\sigma\in \mathbb{S}_N$ and an integer set $A\in [N]$: $\sigma^A$ is a permutation of the elements $[|A|]$ that preserves the relative ordering of $\sigma(i)$, $i\in A$. 
\begin{proposition}\label{lem:dsgeometric}
\cite{li2017efficient} Let $\sigma $ be a permutation generated from the Mallows distribution \eqref{eq:mallow}. Then 
\begin{align*}
    \frac{Pr(\sigma^A(i)=j+1)}{Pr(\sigma^A(i)=j)}\le \max_{\ell \in \{0,\ldots,N-|A|\}}\frac{\phi+\phi^\ell(\sum^{N-\ell-1}_{u=2}\phi^u)}{1+\phi^{2\ell}(\sum^{N-\ell}_{u=3}\phi^u)}
\end{align*}
for $|A|>j\ge \sigma^A_0(i)$ and
\begin{align*}
    \frac{Pr(\sigma^A(i)=j-1)}{Pr(\sigma^A(i)=j)}\le \max_{\ell \in \{0,\ldots,N-|A|\}}\frac{\phi+\phi^\ell(\sum^{N-\ell-1}_{u=2}\phi^u)}{1+\phi^{2\ell}(\sum^{N-\ell}_{u=3}\phi^u)}
\end{align*}
    for $1< j\le \sigma^A_0(i)$.   
\end{proposition}
According to Proposition \ref{lem:dsgeometric}, we have
\begin{align}\label{eq:lehmerdecay}
    \frac{Pr(\sigma^A_{\ell,m}(i)=j)}{Pr(\sigma^A_{\ell,m}(i)=\sigma^A_0(i))}\le p^{|j-\sigma^A_0(i)|},
\end{align}
for any $m\in[m_\ell]$, where $p=\frac{\sum^{N-1}_{u=1}\phi^u}{1+\sum^{N}_{u=3}\phi^u}<1$ when $\phi+\phi^2<1+\phi^N$.

We now show that 
\begin{align}\label{eq:vconcentration}
Pr(v_\ell(i)\ne\mathcal{L}_{\sigma_0}(i))\le e^{-\frac{m_\ell (1-p)^2}{4(1+p)^2}},
\end{align}
where we are using the notation for Lehmer codes from the main text. Note that from \eqref{eq:lehmerdecay}, we have
\begin{align*}
    \frac{Pr(|\sigma^{[i]}(i)-\sigma^{[i]}_0(i)|\ge \frac{\ln (\frac{1}{1-p})}{\ln (\frac{1}{p})})}{Pr(|\sigma^{[i]}(i)-\sigma^{[i]}_0(i)|=1)}\le \frac{p^{\frac{\ln (\frac{1}{1-p})}{\ln (\frac{1}{p})}}}{1-p}=1
\end{align*}
Note that $\sigma^{[i]}_0(i)=i-\mathcal{L}_{\sigma_0}(i)$ and thus that $\mathcal{L}_{\sigma}(i)-\mathcal{L}_{\sigma_0}=\sigma^{[i]}(i)-\sigma^{[i]}_0(i)$. 
Hence,
\begin{align*}
    &Pr(|v_\ell(i)-\mathcal{L}_{\sigma_0}(i)|\ge \frac{\ln (\frac{1}{1-p})}{\ln (\frac{1}{p})} )\\
    \le &Pr(|v_\ell(i)-\mathcal{L}_{\sigma_0}(i)|=1).
\end{align*}
Furthermore, we have
\begin{align}\label{eq:boundonv}
    &Pr(|v_\ell(i)-\mathcal{L}_{\sigma_0}(i)|\ge 1 )\nonumber\\
    \le& \frac{\ln (\frac{1}{1-p})}{\ln (\frac{1}{p})}Pr(|v_\ell(i)-\mathcal{L}_{\sigma_0}(i)|=1).
\end{align}
Next, we compare $Pr(v_\ell(i)=\mathcal{L}_{\sigma_0}(i))$ and $Pr(|v_\ell(i)-\mathcal{L}_{\sigma_0}(i)|=1)$. 
From~\eqref{eq:lehmerdecay}, 
we have that
\begin{align*} 
&Pr(\sigma^{[i]}_{\ell,m}(i)=\sigma^{[i]}_0(i))-Pr(\sigma^{[i]}_{\ell,m}(i)=j)\\
\ge &\frac{1-p^{|j-\sigma^{[i]}_0(i)|}}{1+\frac{2p}{1-p}}\\
=&\frac{(1-p)(1-p^{|j-\sigma^{[i]}_0(i)|})}{(1+p)},
\end{align*}
for $j\ne \sigma^{[i]}_0(i)$. 
According to Heoffding's inequality,
\begin{align}\label{eq:errorforsigman}
    Pr&\Bigg(\Big|\frac{|\{\sigma^{[i]}_{\ell,m}:\sigma^{[i]}_{\ell,m}(i)=\sigma^{[i]}_0(i),m\in [m_\ell]\}|}{m_\ell}\nonumber\\
    &-Pr(\sigma^{[i]}_{\ell,m}(i)=\sigma^{[i]}_0(i))\Big|>\nonumber\\
    &\frac{|Pr(\sigma^{[i]}_{\ell,m}(i)=\sigma^{[i]}_0(i))-Pr(\sigma^{[i]}_{\ell,m}(i)=\sigma^{[i]}_0(i))\pm 1)|}{2}\Bigg)\nonumber\\
    \le &2e^{\frac{-m_\ell(1-p)^4}{2(1+p)^2}}
\end{align}
for any $n\in[N]$. Moreover, 
\begin{align}\label{eq:errorforj}
    Pr&\Bigg(\Big|\frac{|\{\sigma^{[n]}_{\ell,m}:\sigma^{[i]}_{\ell,m}(i)=j,m\in [m_\ell]\}|}{m_\ell}\nonumber\\
    &-Pr(\sigma^{[i]}_{\ell,m}(i)=j)\Big|\nonumber\\
    &>\frac{|Pr(\sigma^{[i]}_{\ell,m}(i)=\sigma^{[i]}_0(i))-Pr(\sigma^{[i]}_{\ell,m}(i)=j|}{2}\Bigg)\nonumber\\
    \le &2e^{\frac{-m_\ell(1-p)^4}{2(1+p)^2}}
\end{align}
for $j= \sigma^{[i]}_0(i)\pm 1$ and $n\in[N]$. Therefore, the probability that $v_\ell(i)\ne \mathcal{L}_{\sigma_0}(i)$ is at most
$$\frac{\ln (\frac{1}{1-p})}{\ln (\frac{1}{p})}Pr(|v_\ell(i)-\mathcal{L}_{\sigma_0}(i)|=1)\le 6\frac{\ln (\frac{1}{1-p})}{\ln (\frac{1}{p})}e^{\frac{-m_\ell(1-p)^4}{2(1+p)^2}},$$
which is at most $e^{-\frac{m_\ell (1-p)^2}{4(1+p)^2}}$ when $m_\ell\ge \frac{4(1+p)^2\ln\Big(6\frac{\ln (\frac{1}{1-p})}{\ln (\frac{1}{p})}\Big)}{(1-p)^2}$. Hence, we have \eqref{eq:vconcentration}. Then, the probability that $v_\ell(i)\ne \mathcal{L}_{\sigma_0}(i)$ for at least half of $\ell\in [L]$ is at most
$$\sum^L_{u = \frac{L}{2}}\binom{L}{u}e^{-u\frac{m_\ell (1-p)^2}{4(1+p)^2}}\le L2^Le^{-\frac{Lm_\ell (1-p)^2}{8(1+p)^2}},$$
which is at most $\frac{\delta}{N}$ when $m_\ell\ge \frac{8(1+p)^2}{(1-p)^2}(\ln 2+\frac{\ln(\frac{NL}{\delta})}{L})$. Therefore, with probability at least $1-\delta$, $v_\ell(i)=\mathcal{L}_{\sigma_0}(i)$ for more than half of the values of $\ell\in[L]$.
\end{proof}
\subsection{Proof of Theorem \ref{thm:lehmercodes}}
\begin{proof} 
From \eqref{eq:lehmerdecay}, we have
\begin{align}\label{eq:dsdecay}
 Pr(\sigma^A_{\ell,m}(i)=j)\le \frac{\epsilon}{MN^2}   
\end{align}
for any $m\in[m_\ell]$, $\ell\in[L]$, $\epsilon>0$, and $$|j-\sigma^A_0(i)|
\ge\frac{\log (\frac{MN^2}{\epsilon})}{\log (\frac{1}{p})}. $$
By the union bound this implies that with probability at least $1-\epsilon$, we have
\begin{align}\label{eq:boundonsigdiff}
    |\sigma^{[i]}_m(i)-\sigma^{[i]}_0(i)|\le \frac{\log (\frac{MN^2}{\epsilon})}{\log (\frac{1}{p})}
\end{align}
for any $m\in[m_\ell]$, $\ell\in[L]$, and $i\in [N]$. Note that $\mathcal{L}_{\sigma}(i)=i-\sigma^{[i]}(i)$ for any permutation $\sigma\in \mathbb{S}_N$ and $i\in [N]$. 
Therefore, to compute the majority of the $i$th coordinate $\mathcal{L}_{\sigma_{\ell,m}}(i)$, $m\in[m_\ell]$, $i\in[N]$, $\ell\in[L]$, it suffices to assume that $\mathcal{L}_{\sigma_{\ell,m}}(i)$, $m\in[m_\ell]$, $i\in[N]$, $\ell\in[L]$,  lies within an interval of length at most $2\frac{\log (\frac{MN^2}{\epsilon})}{\log (\frac{1}{p})}+1$,  
with probability at least $1-\epsilon$. This implies that all $v_\ell(i)$, $\ell\in[L]$, $i\in[N]$, lie in an interval of length at most $I$, where $I=\lceil\log\Big(2\frac{\log (\frac{MN^2}{\epsilon})}{\log (\frac{1}{p})}+1\Big)\rceil$ is defined in \eqref{eq:I} the main text. Hence, the values of the first $\log N-I$ bits of $v_\ell(i)$ differ by at most $1$. Since from Lemma~\ref{lem:majoritylehmer} we know that more than half of the values $v_\ell(i)$ equal $\mathcal{L}_{\sigma_0}(i)$ with probability at least $1-\delta$, averaging the first $\log N-\log I$ bits of $v_\ell(i)$, which produces $\frac{\sum^L_{\ell=1}x^1_\ell(i)}{L}$, gives the first $\log N- \log I$ bits of $\mathcal{L}_{\sigma_0}(i)$. The last $I$ bits of the majority of $v_\ell(i)$, $\ell\in [L]$ can be obtained by identifying the most frequent value in the histogram, computed from the sum of one-hot encodings $\sum^L_{\ell=1}\boldsymbol{x}^2_{\ell,i}$. Therefore, the server recovers $\mathcal{L}_{\sigma_0}$ and thus  $\sigma_0$ with probability at least  $1-\delta-\epsilon$. The communication cost is $\log N-\log I$ for transmitting $x^1_\ell(i)$ and $\log L$ for each coordinate $i$ at each client $L$. Hence the total commmunication cost is $L\Big(N(\log N- \log\Big(2\frac{\log (\frac{MN^2}{\epsilon})}{\log (\frac{1}{p})}+1\Big)+\Big(2\frac{\log (\frac{MN^2}{\epsilon})}{\log (\frac{1}{p})}+1\Big)\log L\Big)$.
\end{proof}
\subsection{Golomb codes}
To compress a random variable $\mathcal{L}_{\sigma_m}(i)$, $i\in[N]$ with geometric distribution $Pr(\mathcal{L}_{\sigma_m}(i)=j)=Pr(\mathcal{L}_{\sigma_m}(i)=0)\phi^j$, the expected number of bits needed is at most $\frac{K'}{1-\phi}+\log K'=O(1)$, where $K'$ is a positive integer satisfying $\phi^{K'}+\phi^{K'+1}\le 1<\phi^{K'-1}+\phi^{K'}$. 

More specifically, for a geometrically distributed random variable $G$ with probability $p=Pr(G=0)$, fix an integer $K$ such that $$(1-p)^K+(1-p)^{K+1}\le 1<(1-p)^{K-1}+(1-p)^K.$$ 
In Golomb codes, $G$ is represented as $G=qK+r$, where $r\in \{0,\ldots,K-1\}$. Note that there is a one-to-one mapping between $G$ and $(q,r)$ where $r\in\{0,\ldots,K-1\}$. Then, we encode $q$ using unary code, which is a string of $q-1$ zero bits followed by a $1$ bit. The remainder $r$ is encoded by its binary presentation of size $\lceil\log_2 K\rceil$. Finally, $G$ is encoded by the concatenation of encodings of $q$ and $r$. When encoding the coordinates $\mathcal{L}_{\sigma_m}(i)$, $i\in[N]$ of Lehmer code using Golomb codes, the expected number of bits is at most 
\begin{align*}
&\frac{(\sum^{K'-1}_{r=0}\phi^r)(\sum^{\lceil\frac{i}{K}\rceil}_{q=1}q\phi^{K'q})}{\sum^{i-1}_{j=0}\phi^j}+\log K'\\
=& \frac{(\sum^{K'-1}_{r=0}\phi^r)(\frac{(\sum^{\lceil\frac{i}{K'}\rceil}_{q=1}\phi^{K'q})-\lceil\frac{\lceil\frac{K'}{i}\rceil\phi^{K'(\lceil\frac{i}{K'}\rceil}}{K'}\rceil+1)}{1-\phi})}{\sum^{i-1}_{j=0}\phi^j}\\
\le& \frac{K'}{1-\phi}+\log K'=O(1),    
\end{align*}
where $\phi^{K'}+\phi^{K'+1}\le 1<\phi^{K'-1}+\phi^{K'}$. Hence, the average communication cost when encoding Lehmer code using Golomb codes is $O(N)$.

However, since Golomb codes are variable-length codes, where the lengths of the codewords depend on their values, it is difficult to apply the secure aggregation scheme because the noise level is predefined before communication. To resolve this issue, one possible approach is to assume that the clients transmit the Golomb codewords of their Lehmer code coordinates in $N$ blocks, each consisting of multiple time slots. At each time slot, a client can either transmit a symbol or remain silent (at no communication cost). The server observes only the sum (over the real field) of the transmitted symbols from clients, without knowing the identity and the value of the clients that transmit the symbols. This may be achieved in a wireless client-server communication setting where the server only receives a superposition of the signals transmitted by the clients. Under this assumption, the server compute the coordinate-wise majority of the Lehmer codes by using Huffman encoding, which is a special case of Golomb codes for $K'=1$ and has similar  communication cost $O(1)$ as shown above. 
Let $\mathcal{H}(\mathcal{L}_{\sigma_m}(i))$ be the Huffman code of the $i$th coordinate of the Lehmer code for permutation $\sigma_m$. 

Let the $i$th, $i\in[N]$ block consist of $i$ time slots. 
At time slot $t$ in the $i$th block, each client $\ell$ transmits the sum (over the real field) of the $t$th bit of $\mathcal{H}(\mathcal{L}_{\sigma_m}(i))$, $m\in\{(\ell-1)c+1,\ldots,\ell c\}$, where $|\mathcal{G}_o(\mathcal{L}_{\sigma_m}(i))|\ge t$. If $|\mathcal{G}_o(\mathcal{L}_{\sigma_m}(i))|< t$ for all $m\in\{(\ell-1)c+1,\ldots,\ell c\}$, client $\ell$ remains silent. The server obtains the number of permutations $\sigma_m$, whose $i$th coordinate of the Lehmer code $\mathcal{L}_{\sigma_m}(i)$ is encoded by at least $t$ bits in the Huffman code, meaning that $\mathcal{L}_{\sigma_m}(i)\ge t-1$. By computing the difference 
\begin{align*}
&|\{\sigma_m:m\in[M],\mathcal{L}_{\sigma_m}(i)\ge t\}|\\
-&|\{\sigma_m:m\in[M],\mathcal{L}_{\sigma_m}(i)\ge t-1\}|    
\end{align*}
for $t\in[i]$
the server obtains the histogram of $\mathcal{L}_{\sigma_m}(i)$, based on which the majority of $\mathcal{L}_{\sigma_m}(i)$ can be computed.


\section{Additonal Experiments}

In what follows, we provide additional experimental results and details regarding our code implementation.

\subsection{Real-world Data}

We evaluate the performance of our FRA algorithms on Sushi preferences, Jester, prioritized cancer gene expression data from The Cancer Genome Atlas (TCGA)~\cite{tcga} and the votes cast during Primary Governor election in Maine in 2018. 

The Sushi dataset consists of $5,000$ rankings of $N=10$ types of sushi. Each ranking has associated metadata indicating the region in Japan where the ranking originated. We convert the ordered lists into complete rankings (or permutations) and filter the data to contain rankings from regions with at least $20$ individuals to arrive at $10$ regions with a total of $M = 4,981$ samples. The only region excluded was foreign which had only $19$ samples.

Each region is treated as a client with access to the rankings belonging to that region only. To study the impact of the number of samples at each client, we use our FRA algorithms to perform aggregation over $200$ subsets of the entire data, with a successively larger subset of samples at each client being used for local aggregation. Since no ground-truth information is available, we evaluate the performance of the methods using the Kemeny objective, normalized by the number of ranked entities. This centralized objective function is the average Kendall $\tau$ distance between $\sigma^*$ and all rankings in the dataset $\sigma_m, m \in [M]$,
$$\frac{1}{MN}\sum^M_{m=1}K_\tau(\sigma^*,\sigma_m).$$

As expected, the performance improves with the number of samples at each client and Borda FRA outperforms other methods (Figure~\ref{fig:real_data_sushi}). Since the number of samples at each client differs substantially, we plot the performance with respect to the minimum number of samples available at any client. For comparison, we also compute an aggregate in a centralized manner assuming that all the rankings are available at the server, and by using the same algorithms. The centralized results are plotted with dashed lines.

The Jester dataset consists of scores in the continuous interval $[-10,10]$ for $N=100$ jokes rated by $48,483$ individuals. We filter the dataset to include $14,116$ individuals who rated all $100$ jokes and converted the scores into complete rankings. The rankings are randomly split into $50$ groups corresponding to the clients. We perform a similar analysis to that described for the Sushi preference dataset and the results are shown in Figure~\ref{fig:real_data_jester}. 

\begin{figure*}[htb]
    \centering
    \begin{subfigure}[t]{0.49\textwidth}
        \centering
        \includegraphics[width=\textwidth]{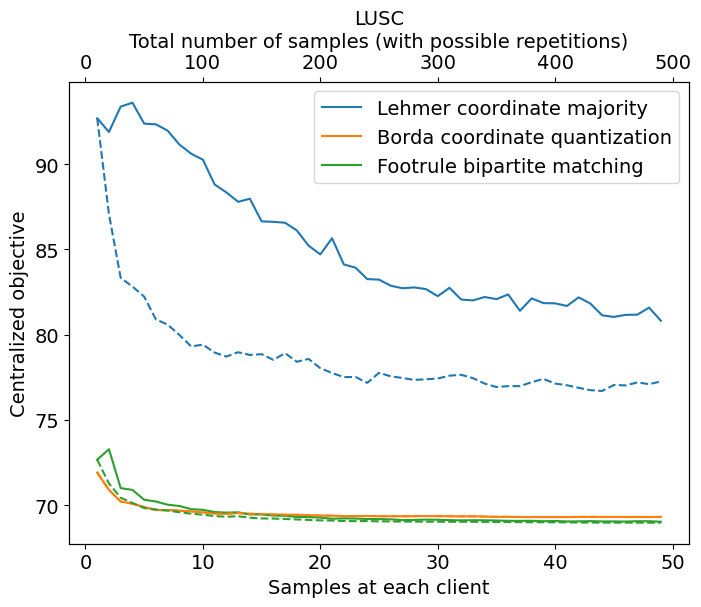} 
        \caption{$N=978, M=491$, $L=10$.} \label{fig:tcga_lusc}
    \end{subfigure}
    \hfill
    \begin{subfigure}[t]{0.49\textwidth}
        \centering
        \includegraphics[width=\textwidth]{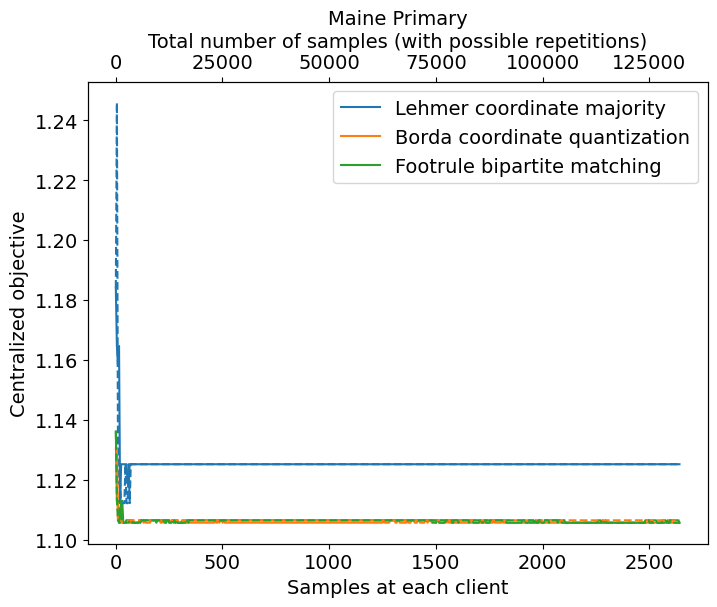} 
        \caption{$N=132,251, M=8$, $L=50$.} \label{fig:maine_gov}
    \end{subfigure}
    \caption{Comparison of the performance of various centralized and FRA methods (centralized results are depicted by dashed, while federated results are indicated by solid lines). (a) Prioritized cancer gene expression data from TCGA for LUSC. (b) Ranked-choice voting data from the primary election for the Governor of Maine in 2018. Here, the performance is evaluated using the Kemeny objective normalized by the permutation length. Results pertaining to the minimum weight bipartite matching aggregation algorithm are shown as well.}
    \label{fig:real_additional}
\end{figure*}

The two prioritized (ranked) gene expression datasets LUAD (lung adenocarcinoma) and LUSC (lung squamous cell carcinoma) from TCGA consist of gene expression profiles of $501$ and $491$ patients, respectively. For prioritization, we select $N =978$ Landmark genes. For each individual, we use the level of expression to arrange the genes into complete rankings. We randomly partition the resulting rankings among $10$ clients. The performance of our algorithms for LUAD is shown in Figure~\ref{fig:real_data_cancer}, while the results for LUSC are shown in Figure~\ref{fig:tcga_lusc}. 

Finally, we tested our algorithms on the ranked-choice voting data for the primary election for the Governor of Maine in 2018. The data consists of $132,251$ ballots. Each ballot has up to $8$ ranked choices that an individual can mark. In ranked-choice voting, there is no restriction on how many choices an individual fills or how they are distributed among the candidates. For example, a person can choose to indicate the same candidate for all $8$ ranks or indicate one candidate for the first $4$ ranks and indicate none for the remaining $4$ ranks.

To convert the ballots to complete rankings, we first use the choices indicated on the ballots to rank as many candidates as possible. Any remaining candidates that were not ranked by the individual are ranked uniformly at random (which is a common practice in social choice theory and practice).

Finally, we randomly assign all the ballots to $L=50$ clients and perform our analysis as described. The results are shown in Figure~\ref{fig:maine_gov}.

\subsection{Synthetic Data}
We tested our FRA algorithms on synthetic data generated using a Mallows model with a wide range of parameter values and federated learning settings.
\begin{enumerate}
    \item The number of clients is fixed as $L=10$, while the number of samples at each client increases $m_l \in \{1,\cdots,50\}, \forall l \in [L]$
    \begin{itemize}
        \item The length of the permutation is $M=10$ (Figure~\ref{fig:synthetic_additional_l10})
        \begin{itemize}
            \item The centroid permutation is the identity permutation $\sigma_0 = e$
            \item The centroid permutation is not the identity permutation $\sigma_0 \neq e$
        \end{itemize}
        \item The length of the permutation is $M=20$ (Figure~\ref{fig:synthetic_additional_l20})
        \begin{itemize}
            \item The centroid permutation is the identity permutation $\sigma_0 = e$
            \item The centroid permutation is not the identity permutation $\sigma_0 \neq e$
        \end{itemize}
    \end{itemize}
    \item The number of samples at each client is fixed as $m_l=10, \forall l \in [L]$, while the number of clients increases $L \in \{1,\cdots,50\}$
    \begin{itemize}
        \item The length of the permutation is $M=10$ (Figure~\ref{fig:synthetic_additional_ml10})
        \begin{itemize}
            \item The centroid permutation is the identity permutation $\sigma_0 = e$
            \item The centroid permutation is not the identity permutation $\sigma_0 \neq e$
        \end{itemize}
        \item The length of the permutation is $M=20$ (Figure~\ref{fig:synthetic_additional_ml20})
        \begin{itemize}
            \item The centroid permutation is the identity permutation $\sigma_0 = e$
            \item The centroid permutation is not the identity permutation $\sigma_0 \neq e$
        \end{itemize}
    \end{itemize}
\end{enumerate}

For each of the above combinations of parameters, we repeat the experiments for scaling factor $\phi \in \{0.1,0.2,0.3,0.4,0.5,0.6\}$.

\begin{figure*}[htb]
    \centering
    \begin{subfigure}[t]{0.49\textwidth}
        \centering
        \includegraphics[width=\textwidth]{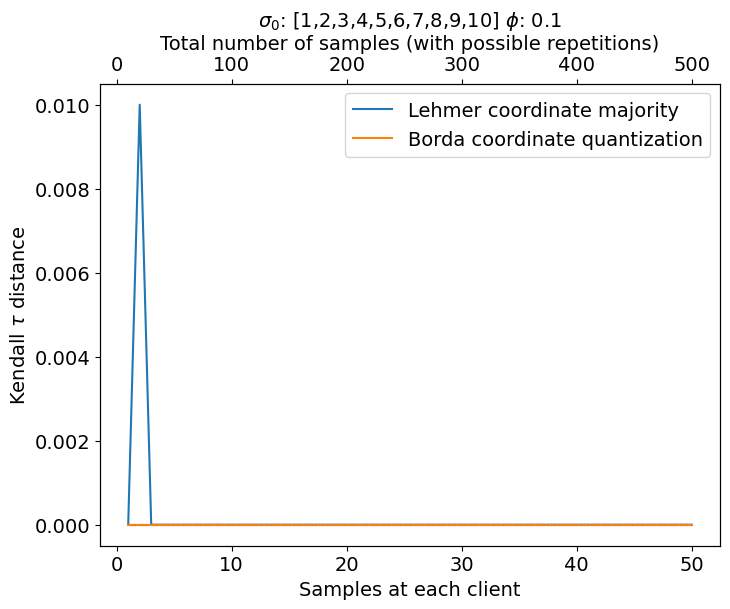} 
        \caption{$\sigma_0 = e, \phi=0.1$, $L=10$.} \label{}
    \end{subfigure}
    \hfill
    \begin{subfigure}[t]{0.49\textwidth}
        \centering
        \includegraphics[width=\textwidth]{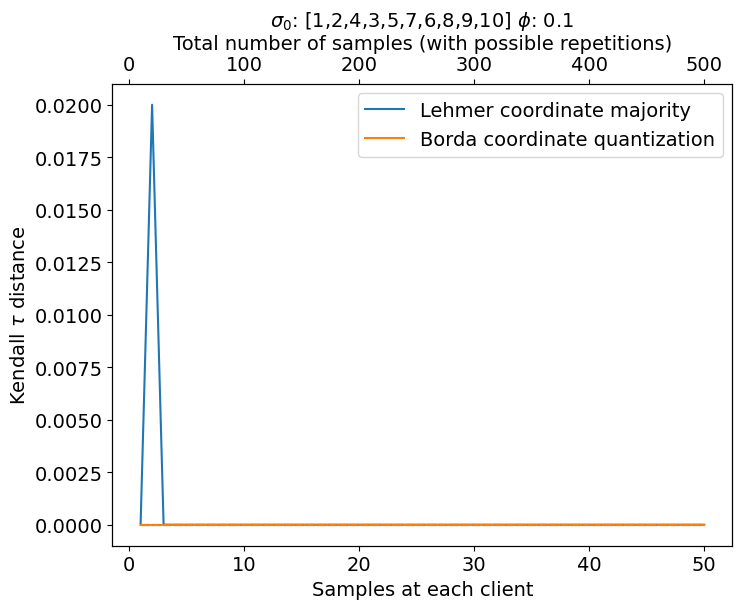} 
        \caption{$\sigma_0 \neq e, \phi=0.1$, $L=10$.} \label{}
    \end{subfigure}
    \begin{subfigure}[t]{0.49\textwidth}
        \centering
        \includegraphics[width=\textwidth]{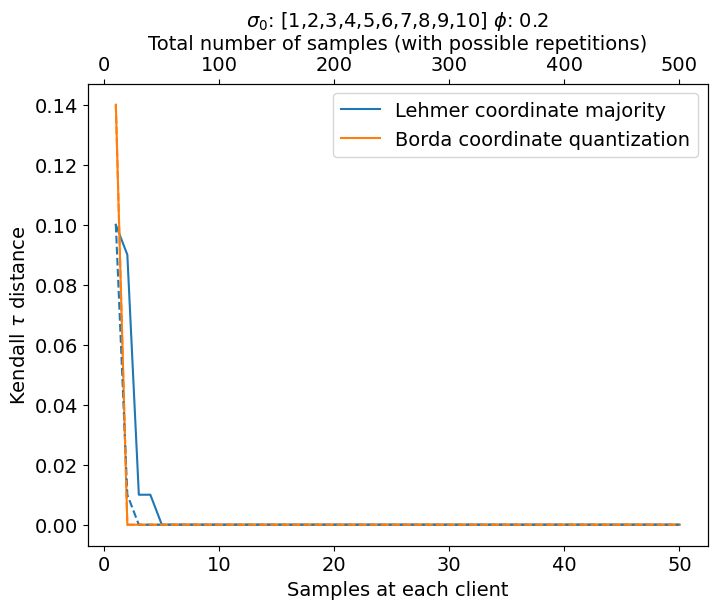} 
        \caption{$\sigma_0 = e, \phi=0.2$, $L=10$.} \label{}
    \end{subfigure}
    \hfill
    \begin{subfigure}[t]{0.49\textwidth}
        \centering
        \includegraphics[width=\textwidth]{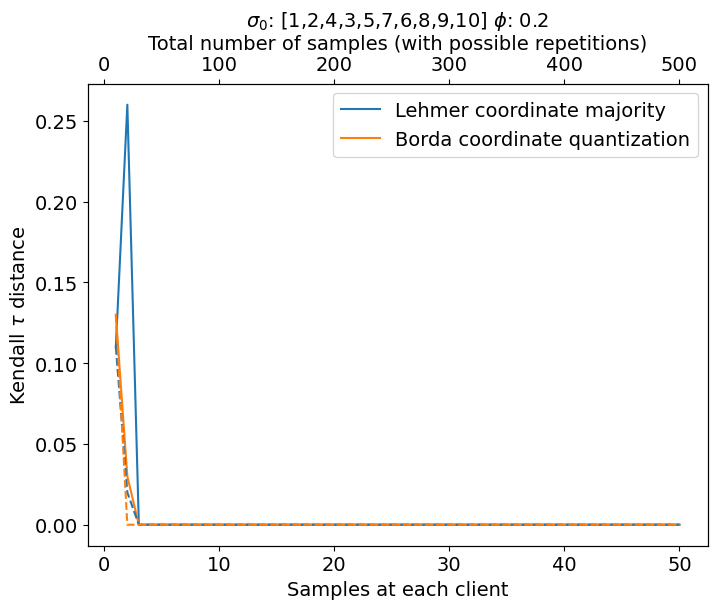} 
        \caption{$\sigma_0 \neq e, \phi=0.2$, $L=10$.} \label{}
    \end{subfigure}
    \begin{subfigure}[t]{0.49\textwidth}
        \centering
        \includegraphics[width=\textwidth]{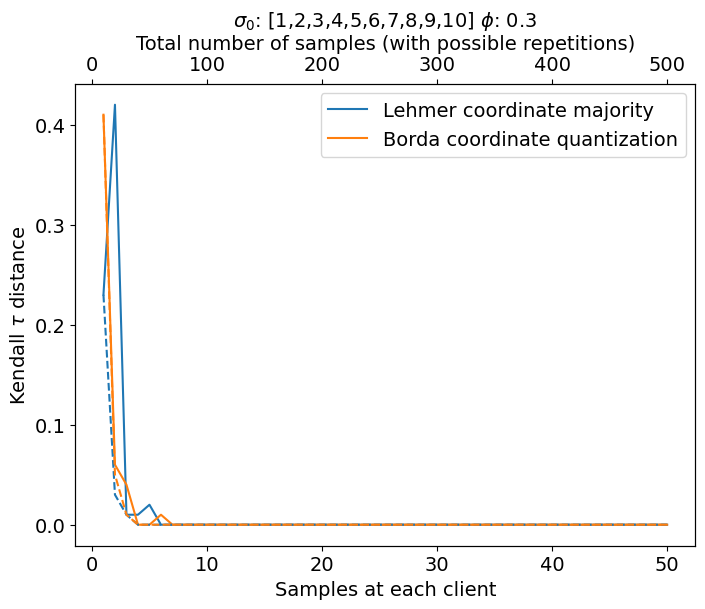} 
        \caption{$\sigma_0 = e, \phi=0.3$, $L=10$.} \label{}
    \end{subfigure}
    \hfill
    \begin{subfigure}[t]{0.49\textwidth}
        \centering
        \includegraphics[width=\textwidth]{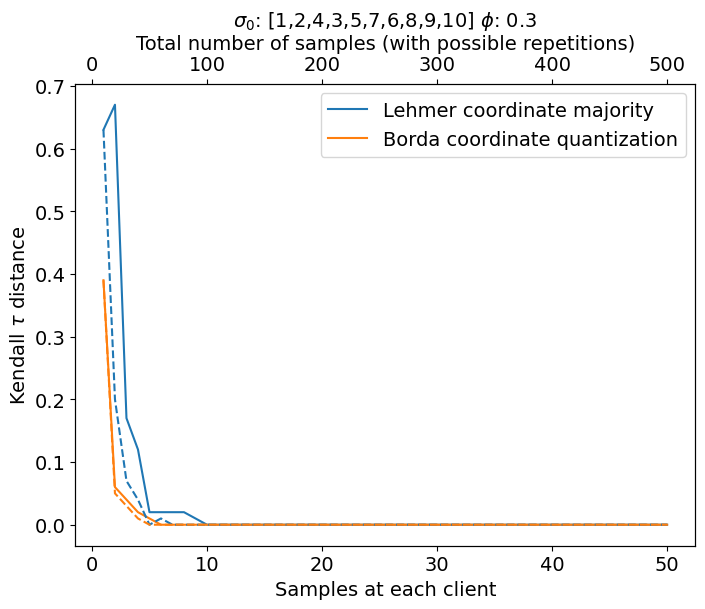} 
        \caption{$\sigma_0 \neq e, \phi=0.3$, $L=10$.} \label{}
    \end{subfigure}

    \caption{Comparison of the performance of various centralized and FRA methods (centralized results are depicted by dashed, while federated results are indicated by solid lines). Plots show the average Kendall $\tau$ distances between $\sigma_0$ and its estimate, for different values of $\phi$.}
    \label{fig:synthetic_additional_l10}
\end{figure*}

\begin{figure*}[htb]
\ContinuedFloat
    \centering
    \begin{subfigure}[t]{0.49\textwidth}
        \centering
        \includegraphics[width=\textwidth]{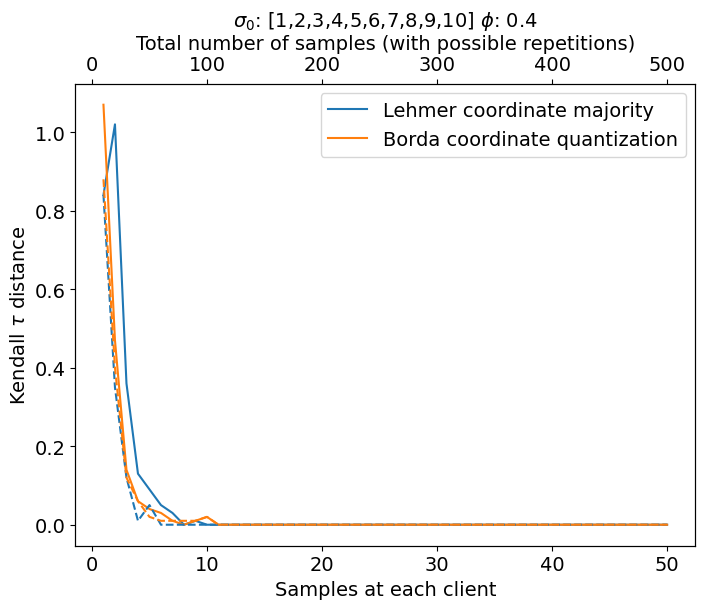} 
        \caption{$\sigma_0 = e, \phi=0.4$, $L=10$.} \label{}
    \end{subfigure}
    \hfill
    \begin{subfigure}[t]{0.49\textwidth}
        \centering
        \includegraphics[width=\textwidth]{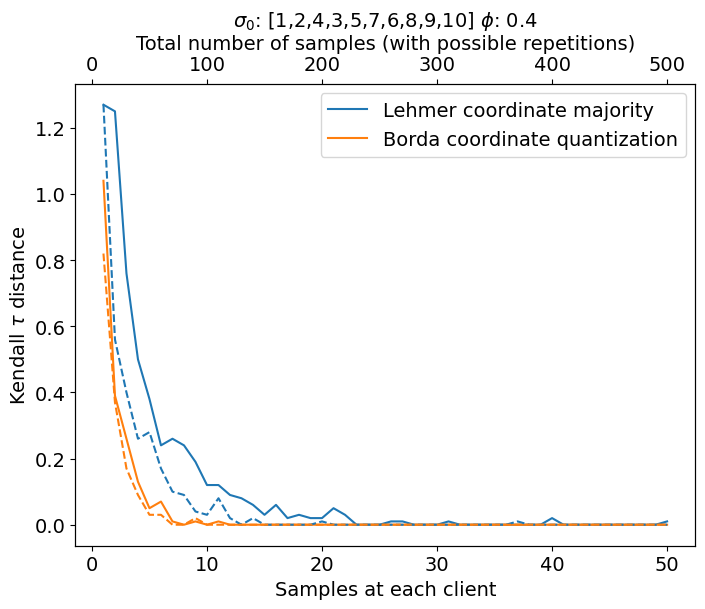} 
        \caption{$\sigma_0 \neq e, \phi=0.4$, $L=10$.} \label{}
    \end{subfigure}
    \begin{subfigure}[t]{0.49\textwidth}
        \centering
        \includegraphics[width=\textwidth]{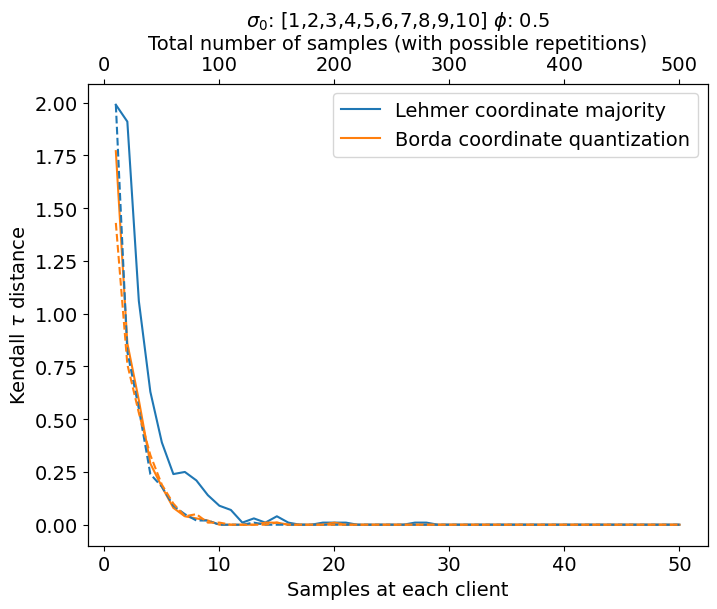} 
        \caption{$\sigma_0 = e, \phi=0.5$, $L=10$.} \label{}
    \end{subfigure}
    \hfill
    \begin{subfigure}[t]{0.49\textwidth}
        \centering
        \includegraphics[width=\textwidth]{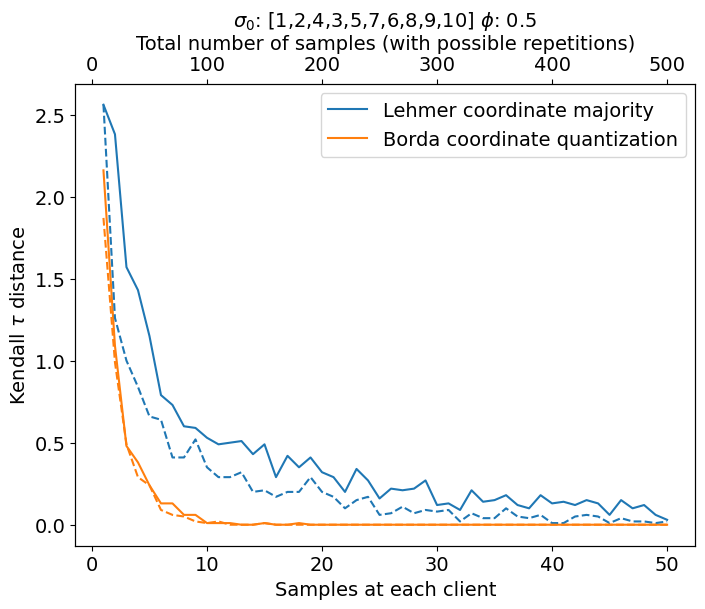} 
        \caption{$\sigma_0 \neq e, \phi=0.5$, $L=10$.} \label{}
    \end{subfigure}
    \begin{subfigure}[t]{0.49\textwidth}
        \centering
        \includegraphics[width=\textwidth]{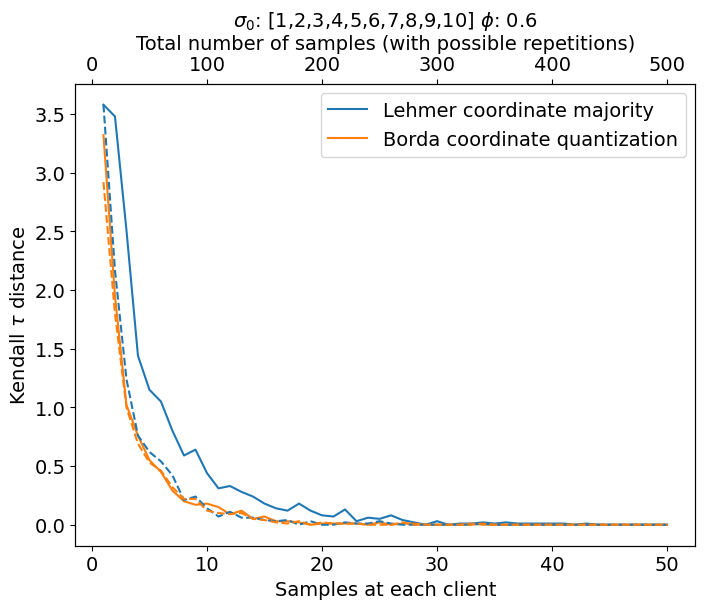} 
        \caption{$\sigma_0 = e, \phi=0.6$, $L=10$.} \label{}
    \end{subfigure}
    \hfill
    \begin{subfigure}[t]{0.49\textwidth}
        \centering
        \includegraphics[width=\textwidth]{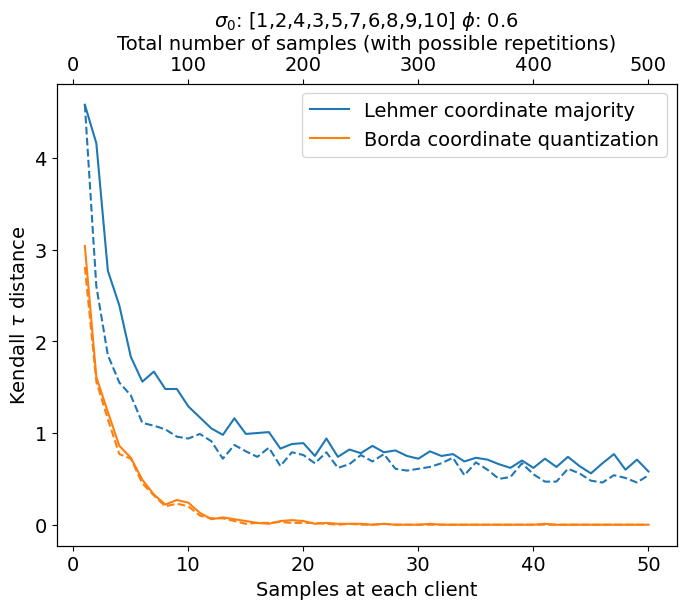} 
        \caption{$\sigma_0 \neq e, \phi=0.6$, $L=10$.} \label{}
    \end{subfigure}

    \caption{Comparison of the performance of various centralized and FRA methods (centralized results are depicted by dashed, while federated results are indicated by solid lines). Plots show the average Kendall $\tau$ distances between $\sigma_0$ and its estimate, for different values of $\phi$.}
\end{figure*}


\begin{figure*}[htb]
    \centering
    \begin{subfigure}[t]{0.49\textwidth}
        \centering
        \includegraphics[width=\textwidth]{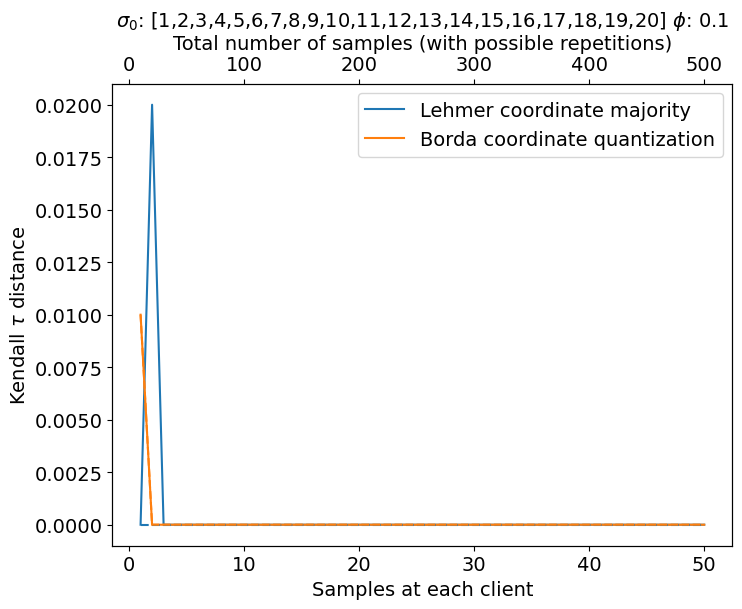} 
        \caption{$\sigma_0 = e, \phi=0.1$, $L=10$.} \label{}
    \end{subfigure}
    \hfill
    \begin{subfigure}[t]{0.49\textwidth}
        \centering
        \includegraphics[width=\textwidth]{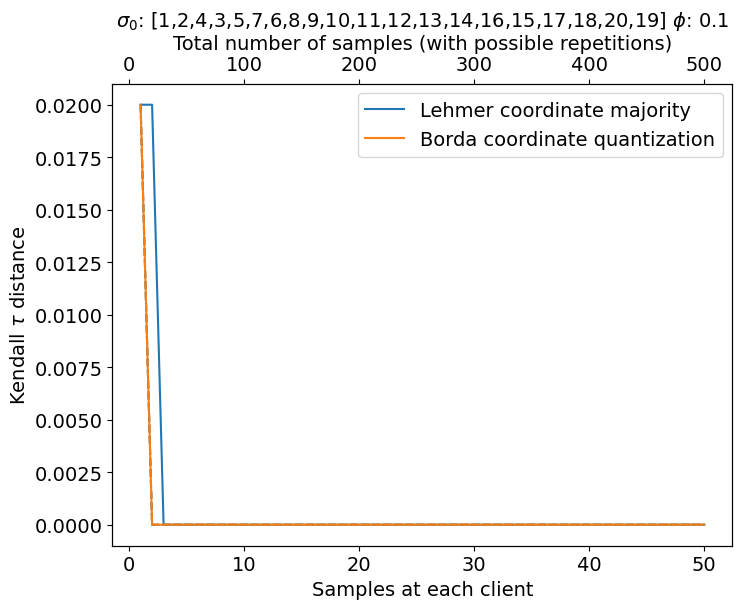} 
        \caption{$\sigma_0 \neq e, \phi=0.1$, $L=10$.} \label{}
    \end{subfigure}
    \begin{subfigure}[t]{0.49\textwidth}
        \centering
        \includegraphics[width=\textwidth]{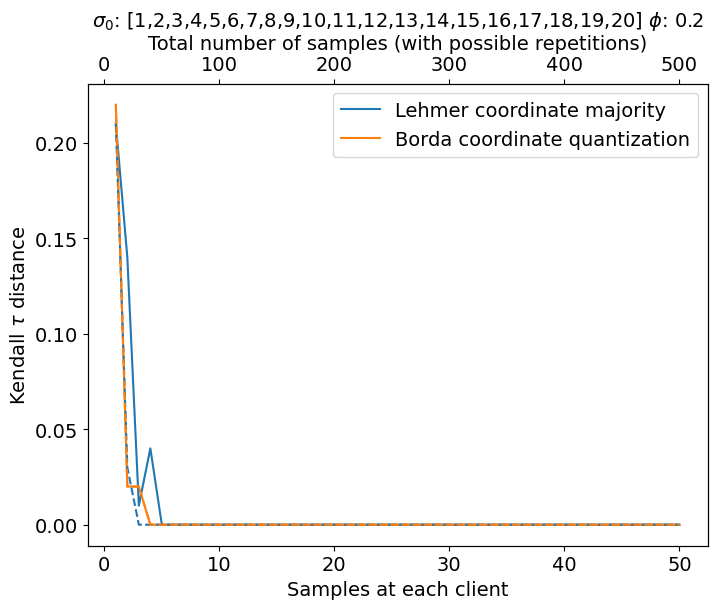} 
        \caption{$\sigma_0 = e, \phi=0.2$, $L=10$.} \label{}
    \end{subfigure}
    \hfill
    \begin{subfigure}[t]{0.49\textwidth}
        \centering
        \includegraphics[width=\textwidth]{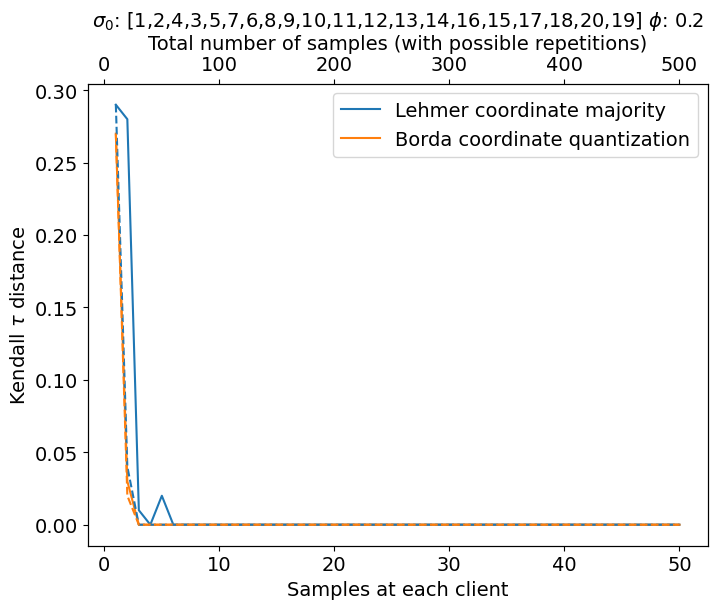} 
        \caption{$\sigma_0 \neq e, \phi=0.2$, $L=10$.} \label{}
    \end{subfigure}
    \begin{subfigure}[t]{0.49\textwidth}
        \centering
        \includegraphics[width=\textwidth]{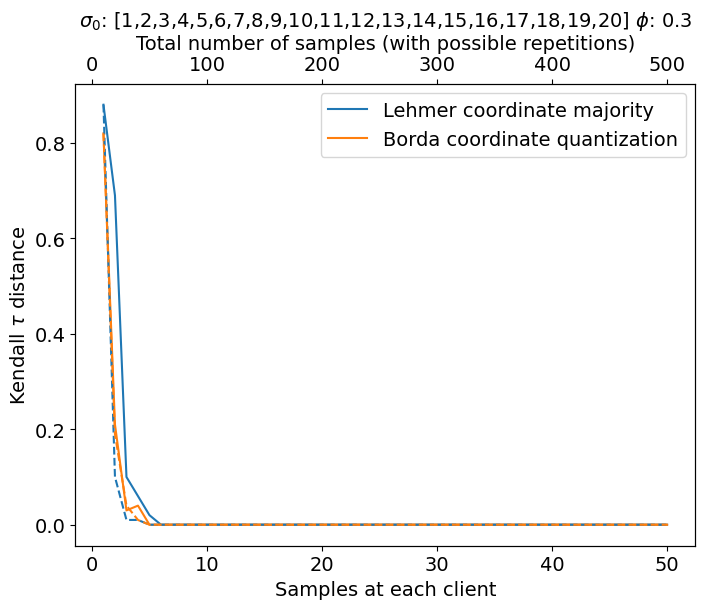} 
        \caption{$\sigma_0 = e, \phi=0.3$, $L=10$.} \label{}
    \end{subfigure}
    \hfill
    \begin{subfigure}[t]{0.49\textwidth}
        \centering
        \includegraphics[width=\textwidth]{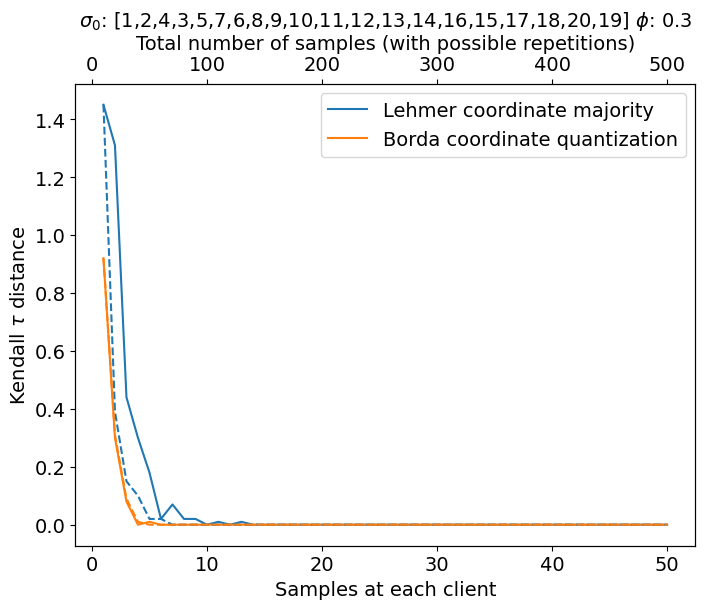} 
        \caption{$\sigma_0 \neq e, \phi=0.3$, $L=10$.} \label{}
    \end{subfigure}

    \caption{Comparison of the performance of various centralized and FRA methods (centralized results are depicted by dashed, while federated results are indicated by solid lines). Plots show the average Kendall $\tau$ distances between $\sigma_0$ and its estimate, for different values of $\phi$.}
    \label{fig:synthetic_additional_l20}
\end{figure*}

\begin{figure*}[htb]
\ContinuedFloat
    \centering
    \begin{subfigure}[t]{0.49\textwidth}
        \centering
        \includegraphics[width=\textwidth]{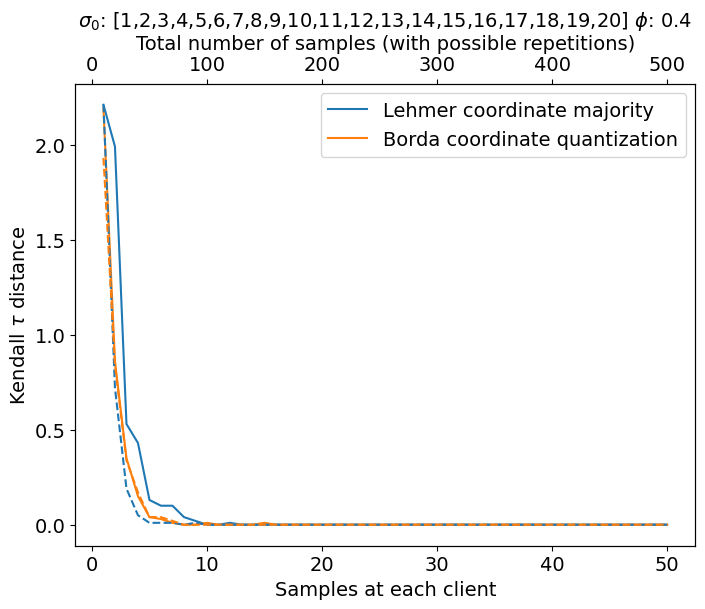} 
        \caption{$\sigma_0 = e, \phi=0.4$, $L=10$.} \label{}
    \end{subfigure}
    \hfill
    \begin{subfigure}[t]{0.49\textwidth}
        \centering
        \includegraphics[width=\textwidth]{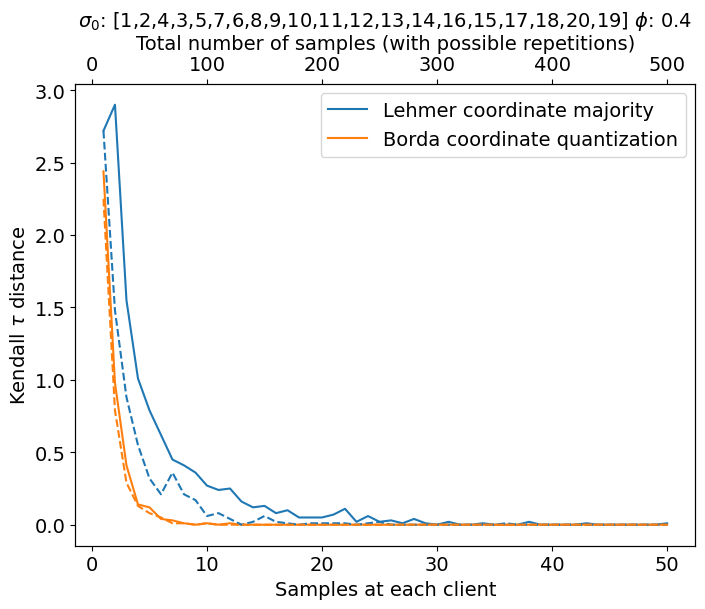} 
        \caption{$\sigma_0 \neq e, \phi=0.4$, $L=10$.} \label{}
    \end{subfigure}
    \begin{subfigure}[t]{0.49\textwidth}
        \centering
        \includegraphics[width=\textwidth]{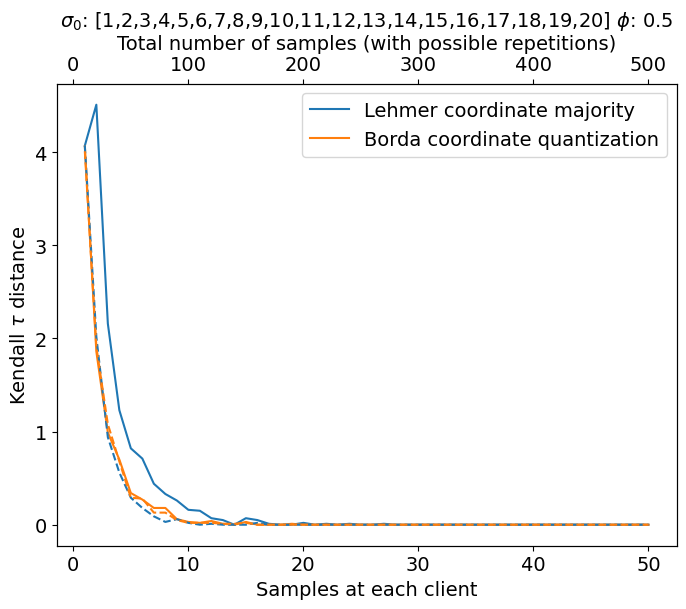} 
        \caption{$\sigma_0 = e, \phi=0.5$, $L=10$.} \label{}
    \end{subfigure}
    \hfill
    \begin{subfigure}[t]{0.49\textwidth}
        \centering
        \includegraphics[width=\textwidth]{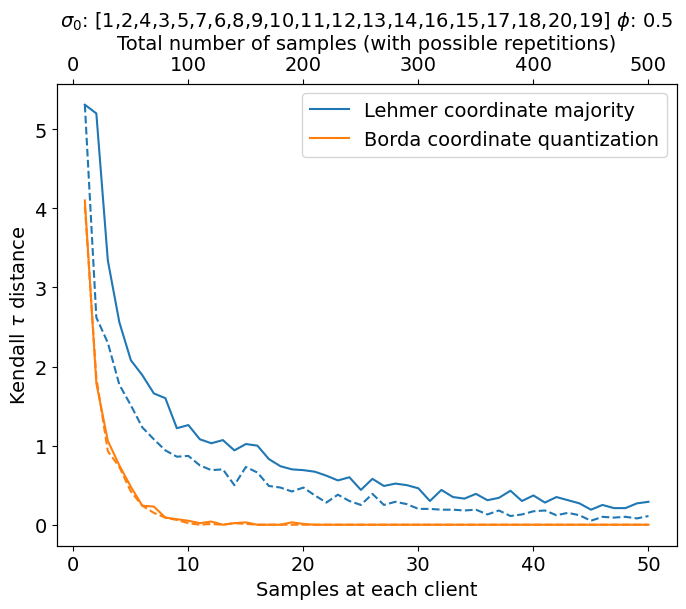} 
        \caption{$\sigma_0 \neq e, \phi=0.5$, $L=10$.} \label{}
    \end{subfigure}
    \begin{subfigure}[t]{0.49\textwidth}
        \centering
        \includegraphics[width=\textwidth]{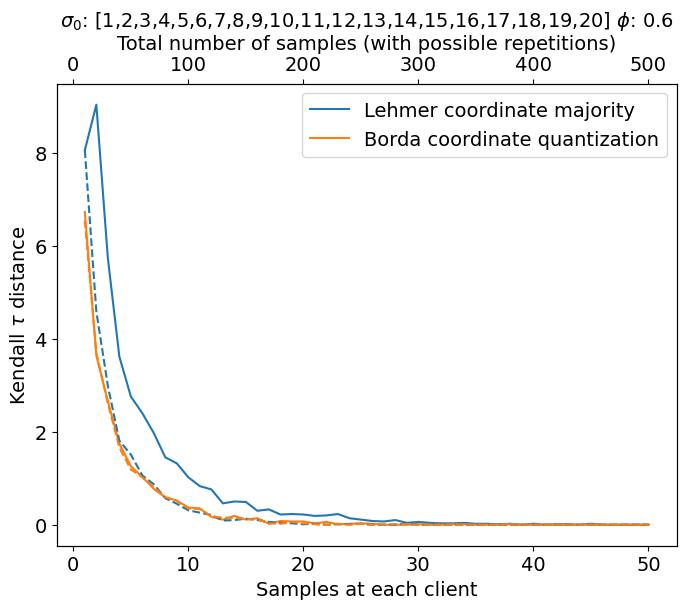} 
        \caption{$\sigma_0 = e, \phi=0.6$, $L=10$.} \label{}
    \end{subfigure}
    \hfill
    \begin{subfigure}[t]{0.49\textwidth}
        \centering
        \includegraphics[width=\textwidth]{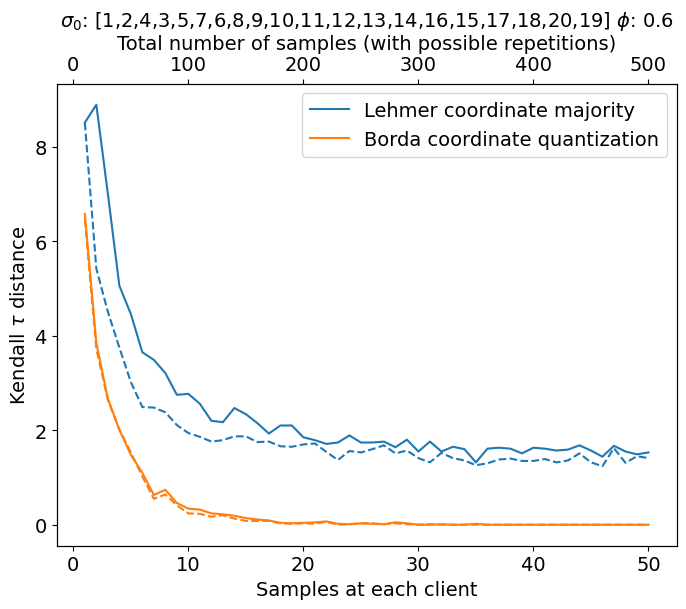} 
        \caption{$\sigma_0 \neq e, \phi=0.6$, $L=10$.} \label{}
    \end{subfigure}

    \caption{Comparison of the performance of various centralized and FRA methods (centralized results are depicted by dashed, while federated results are indicated by solid lines). Plots show the average Kendall $\tau$ distances between $\sigma_0$ and its estimate, for different values of $\phi$.}
\end{figure*}


\begin{figure*}[htb]
    \centering
    \begin{subfigure}[t]{0.49\textwidth}
        \centering
        \includegraphics[width=\textwidth]{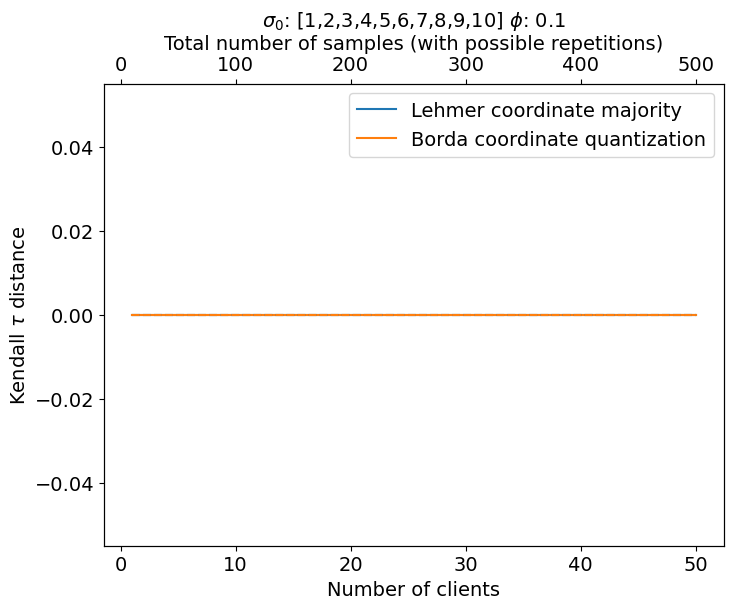} 
        \caption{$\sigma_0 = e, \phi=0.1$, $m_l=10$.} \label{}
    \end{subfigure}
    \hfill
    \begin{subfigure}[t]{0.49\textwidth}
        \centering
        \includegraphics[width=\textwidth]{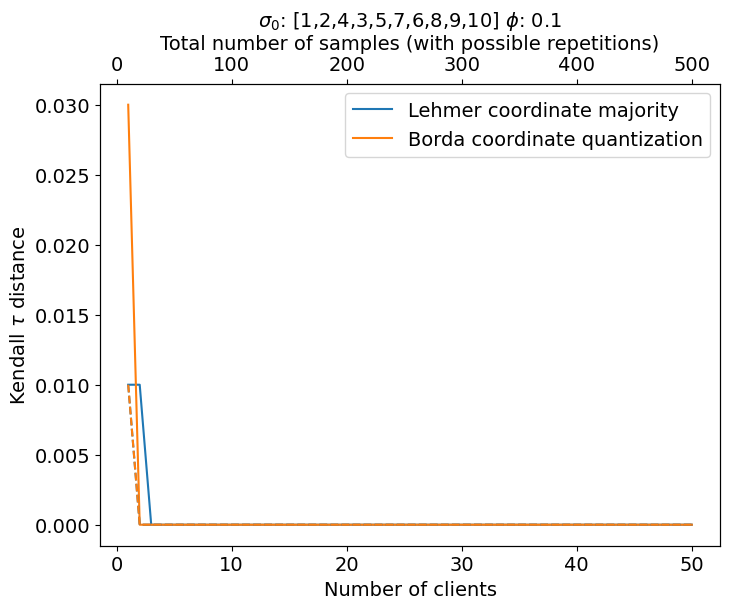} 
        \caption{$\sigma_0 \neq e, \phi=0.1$, $m_l=10$.} \label{}
    \end{subfigure}
    \begin{subfigure}[t]{0.49\textwidth}
        \centering
        \includegraphics[width=\textwidth]{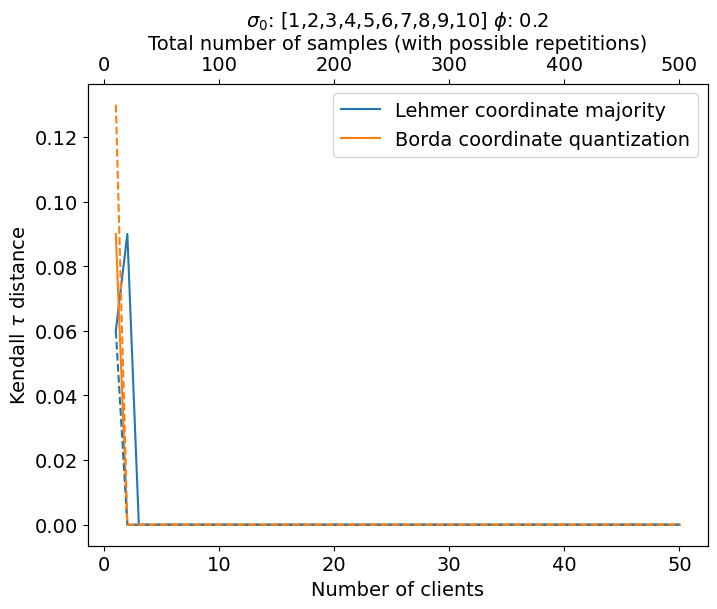} 
        \caption{$\sigma_0 = e, \phi=0.2$, $m_l=10$.} \label{}
    \end{subfigure}
    \hfill
    \begin{subfigure}[t]{0.49\textwidth}
        \centering
        \includegraphics[width=\textwidth]{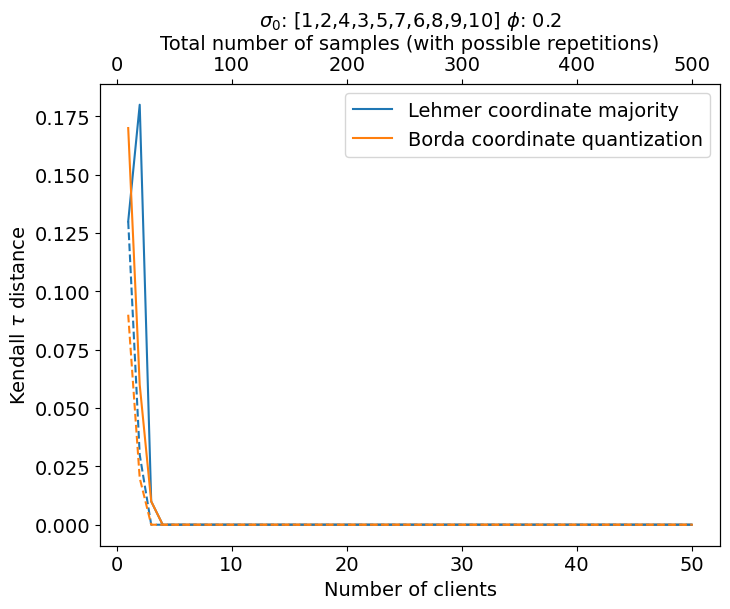} 
        \caption{$\sigma_0 \neq e, \phi=0.2$, $m_l=10$.} \label{}
    \end{subfigure}
    \begin{subfigure}[t]{0.49\textwidth}
        \centering
        \includegraphics[width=\textwidth]{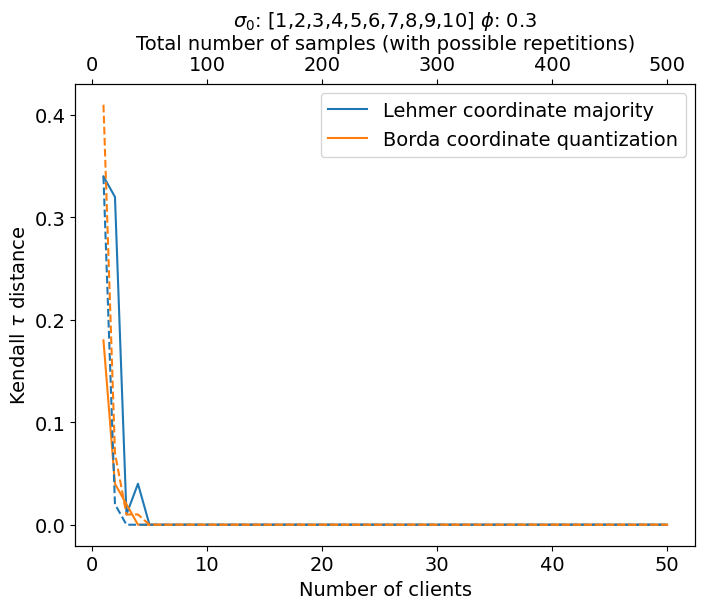} 
        \caption{$\sigma_0 = e, \phi=0.3$, $m_l=10$.} \label{}
    \end{subfigure}
    \hfill
    \begin{subfigure}[t]{0.49\textwidth}
        \centering
        \includegraphics[width=\textwidth]{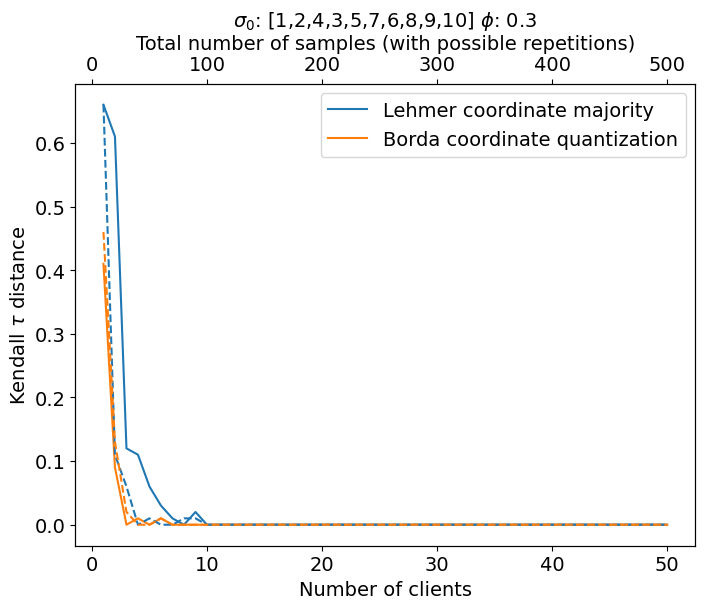} 
        \caption{$\sigma_0 \neq e, \phi=0.3$, $m_l=10$.} \label{}
    \end{subfigure}

    \caption{Comparison of the performance of various centralized and FRA methods (centralized results are depicted by dashed, while federated results are indicated by solid lines). Plots show the average Kendall $\tau$ distances between $\sigma_0$ and its estimate, for different values of $\phi$.}
    \label{fig:synthetic_additional_ml10}
\end{figure*}

\begin{figure*}[htb]
\ContinuedFloat
    \centering
    \begin{subfigure}[t]{0.49\textwidth}
        \centering
        \includegraphics[width=\textwidth]{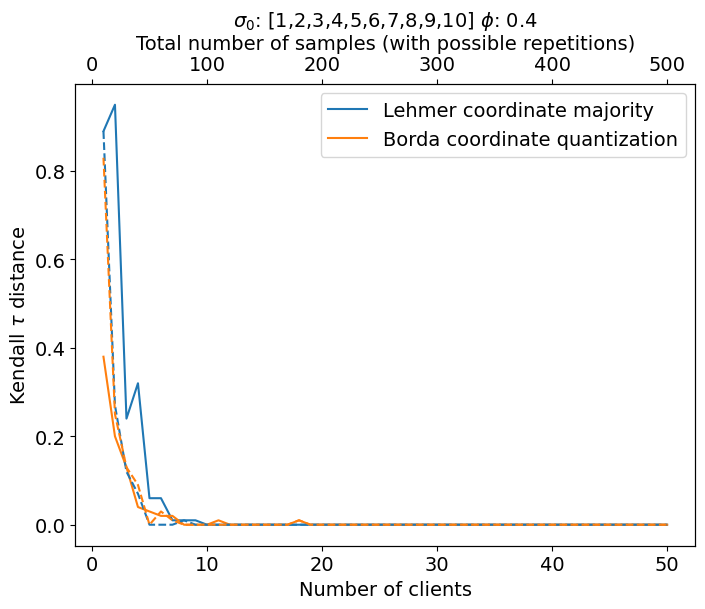} 
        \caption{$\sigma_0 = e, \phi=0.4$, $m_l=10$.} \label{}
    \end{subfigure}
    \hfill
    \begin{subfigure}[t]{0.49\textwidth}
        \centering
        \includegraphics[width=\textwidth]{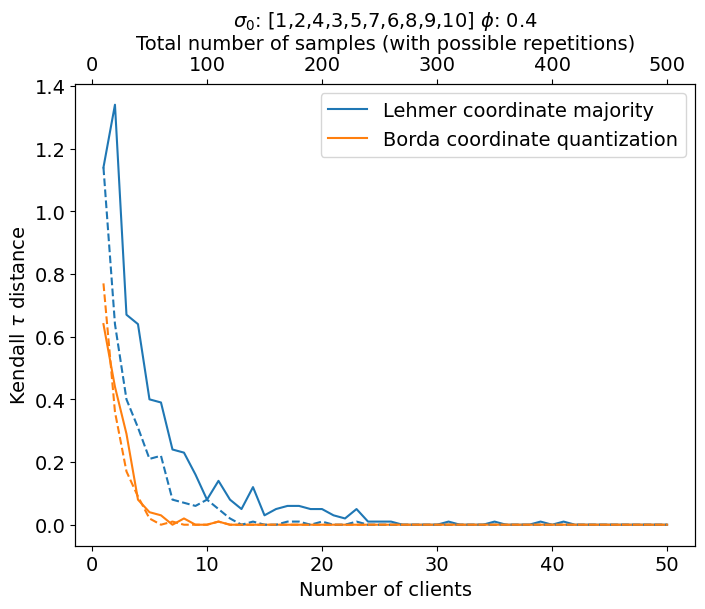} 
        \caption{$\sigma_0 \neq e, \phi=0.4$, $m_l=10$.} \label{}
    \end{subfigure}
    \begin{subfigure}[t]{0.49\textwidth}
        \centering
        \includegraphics[width=\textwidth]{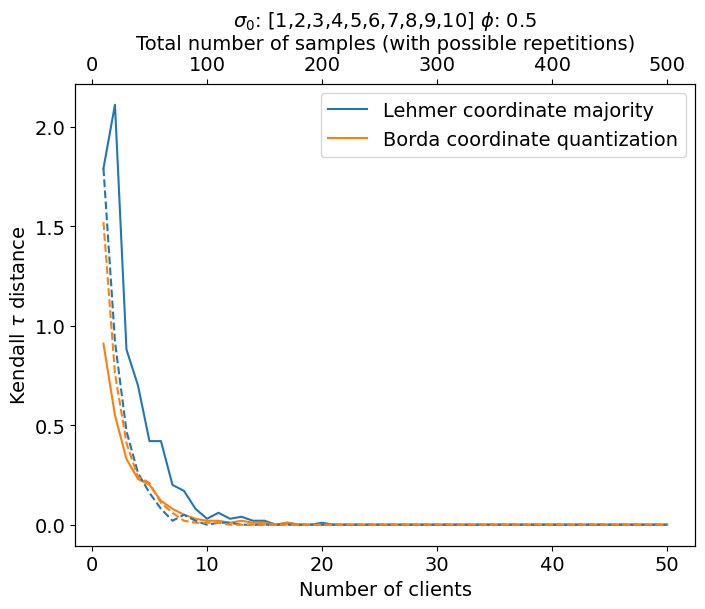} 
        \caption{$\sigma_0 = e, \phi=0.5$, $m_l=10$.} \label{}
    \end{subfigure}
    \hfill
    \begin{subfigure}[t]{0.49\textwidth}
        \centering
        \includegraphics[width=\textwidth]{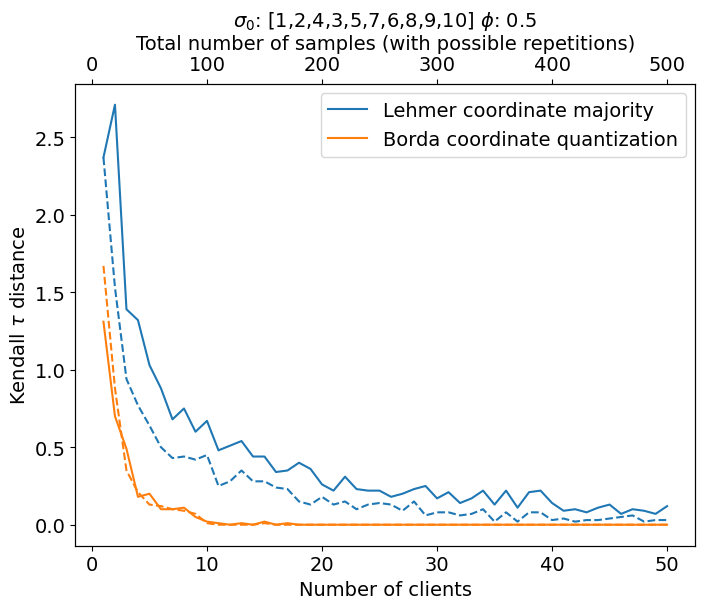} 
        \caption{$\sigma_0 \neq e, \phi=0.5$, $m_l=10$.} \label{}
    \end{subfigure}
    \begin{subfigure}[t]{0.49\textwidth}
        \centering
        \includegraphics[width=\textwidth]{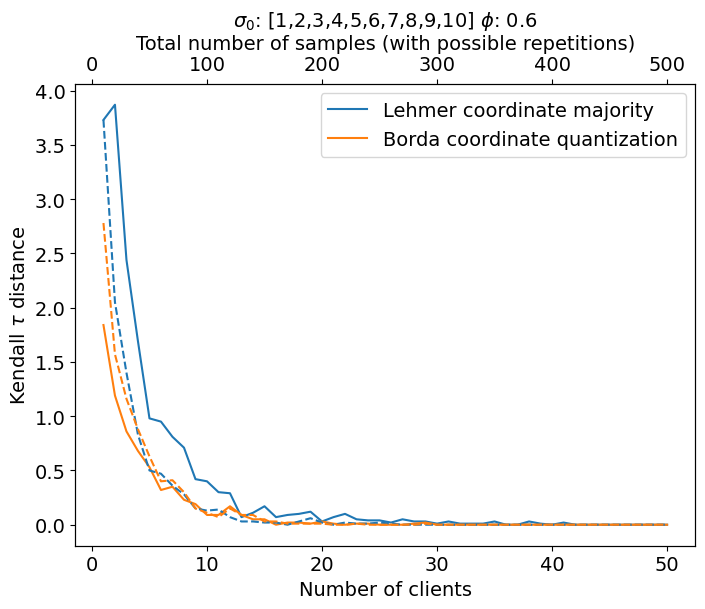} 
        \caption{$\sigma_0 = e, \phi=0.6$, $m_l=10$.} \label{}
    \end{subfigure}
    \hfill
    \begin{subfigure}[t]{0.49\textwidth}
        \centering
        \includegraphics[width=\textwidth]{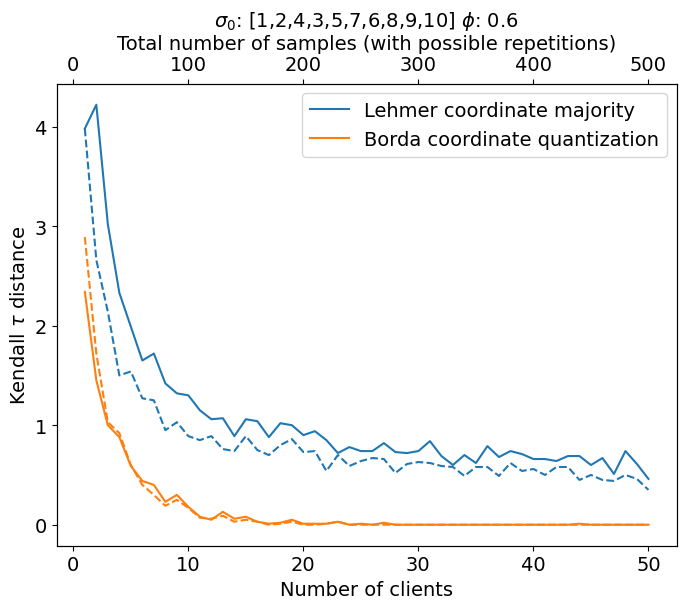} 
        \caption{$\sigma_0 \neq e, \phi=0.6$, $m_l=10$.} \label{}
    \end{subfigure}

    \caption{Comparison of the performance of various centralized and FRA methods (centralized results are depicted by dashed, while federated results are indicated by solid lines). Plots show the average Kendall $\tau$ distances between $\sigma_0$ and its estimate, for different values of $\phi$.}
\end{figure*}


\begin{figure*}[htb]
    \centering
    \begin{subfigure}[t]{0.49\textwidth}
        \centering
        \includegraphics[width=\textwidth]{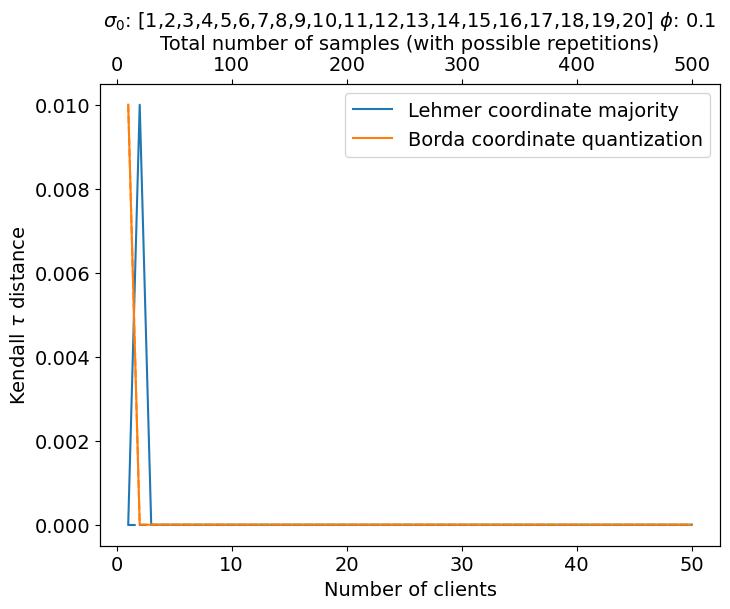} 
        \caption{$\sigma_0 = e, \phi=0.1$, $m_l=10$.} \label{}
    \end{subfigure}
    \hfill
    \begin{subfigure}[t]{0.49\textwidth}
        \centering
        \includegraphics[width=\textwidth]{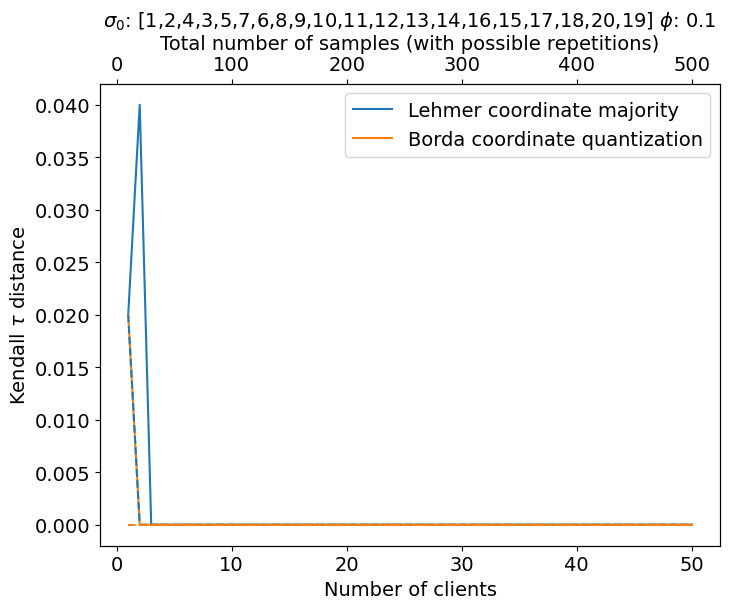} 
        \caption{$\sigma_0 \neq e, \phi=0.1$, $m_l=10$.} \label{}
    \end{subfigure}
    \begin{subfigure}[t]{0.49\textwidth}
        \centering
        \includegraphics[width=\textwidth]{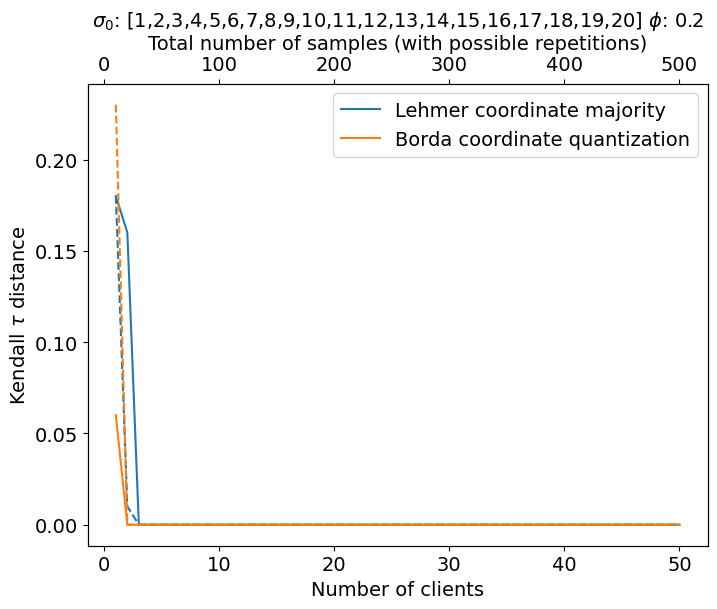} 
        \caption{$\sigma_0 = e, \phi=0.2$, $m_l=10$.} \label{}
    \end{subfigure}
    \hfill
    \begin{subfigure}[t]{0.49\textwidth}
        \centering
        \includegraphics[width=\textwidth]{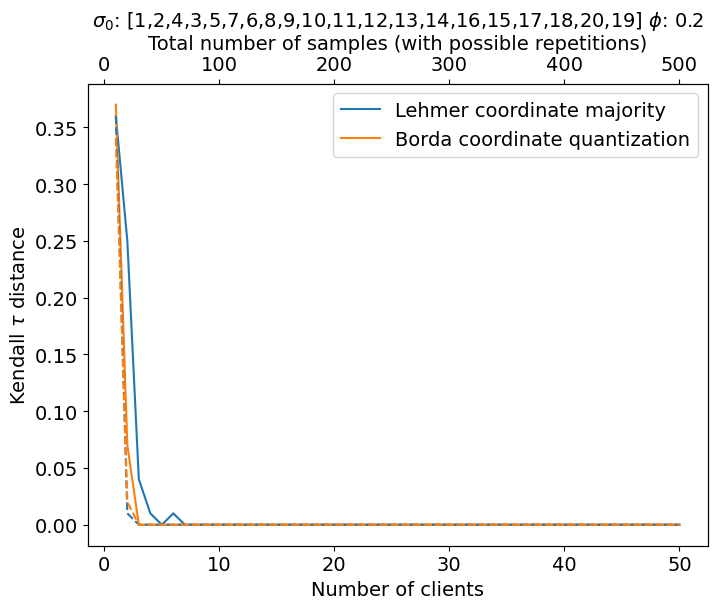} 
        \caption{$\sigma_0 \neq e, \phi=0.2$, $m_l=10$.} \label{}
    \end{subfigure}
    \begin{subfigure}[t]{0.49\textwidth}
        \centering
        \includegraphics[width=\textwidth]{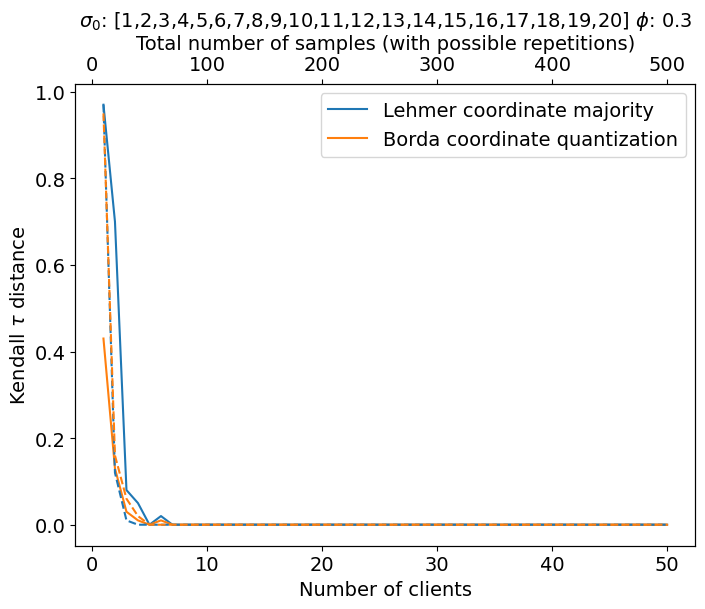} 
        \caption{$\sigma_0 = e, \phi=0.3$, $m_l=10$.} \label{}
    \end{subfigure}
    \hfill
    \begin{subfigure}[t]{0.49\textwidth}
        \centering
        \includegraphics[width=\textwidth]{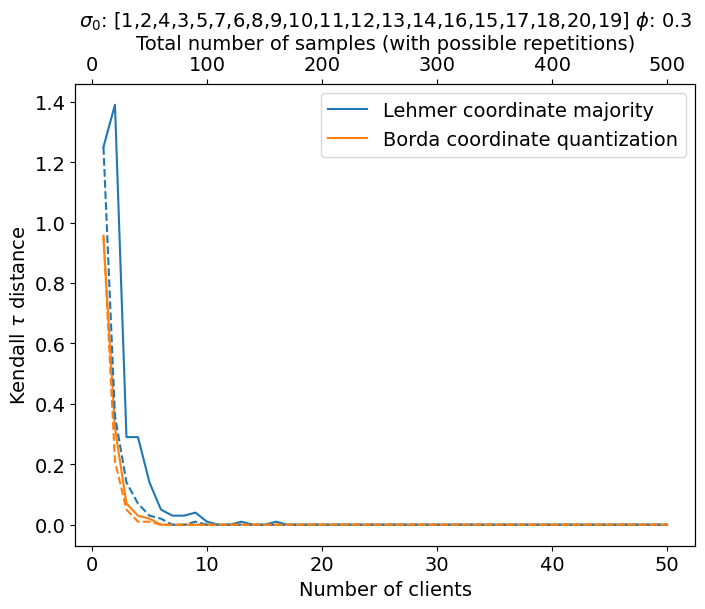} 
        \caption{$\sigma_0 \neq e, \phi=0.3$, $m_l=10$.} \label{}
    \end{subfigure}

    \caption{Comparison of the performance of various centralized and FRA methods (centralized results are depicted by dashed, while federated results are indicated by solid lines). Plots show the average Kendall $\tau$ distances between $\sigma_0$ and its estimate, for different values of $\phi$.}
    \label{fig:synthetic_additional_ml20}
\end{figure*}

\begin{figure*}[htb]
\ContinuedFloat
    \centering
    \begin{subfigure}[t]{0.49\textwidth}
        \centering
        \includegraphics[width=\textwidth]{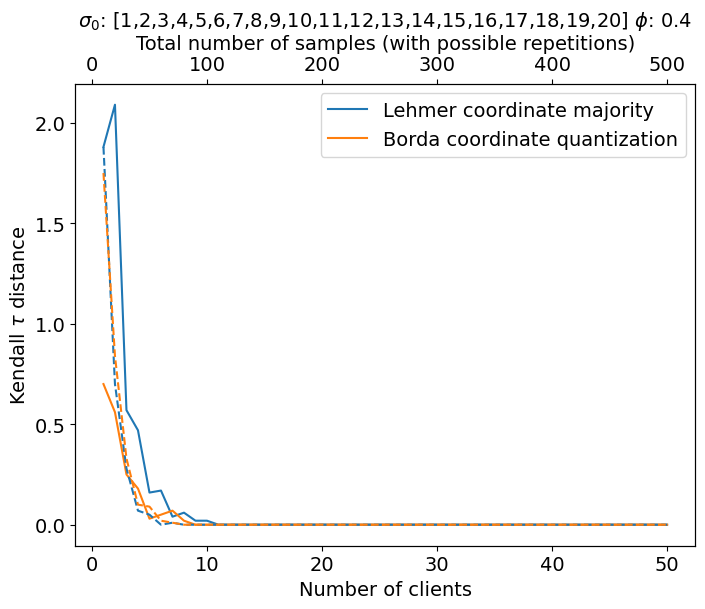} 
        \caption{$\sigma_0 = e, \phi=0.4$, $m_l=10$.} \label{}
    \end{subfigure}
    \hfill
    \begin{subfigure}[t]{0.49\textwidth}
        \centering
        \includegraphics[width=\textwidth]{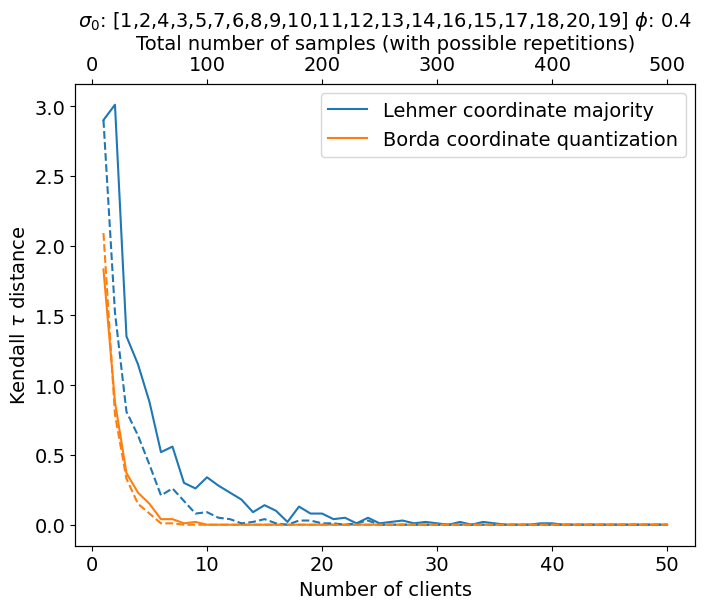} 
        \caption{$\sigma_0 \neq e, \phi=0.4$, $m_l=10$.} \label{}
    \end{subfigure}
    \begin{subfigure}[t]{0.49\textwidth}
        \centering
        \includegraphics[width=\textwidth]{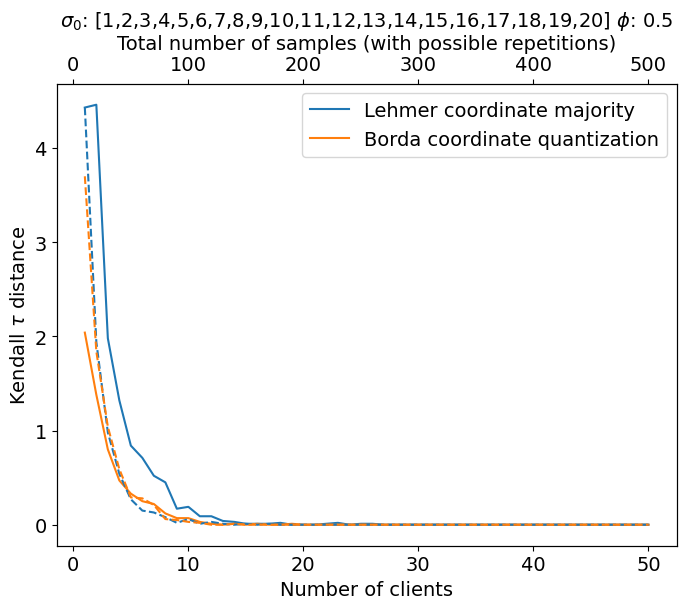} 
        \caption{$\sigma_0 = e, \phi=0.5$, $m_l=10$.} \label{}
    \end{subfigure}
    \hfill
    \begin{subfigure}[t]{0.49\textwidth}
        \centering
        \includegraphics[width=\textwidth]{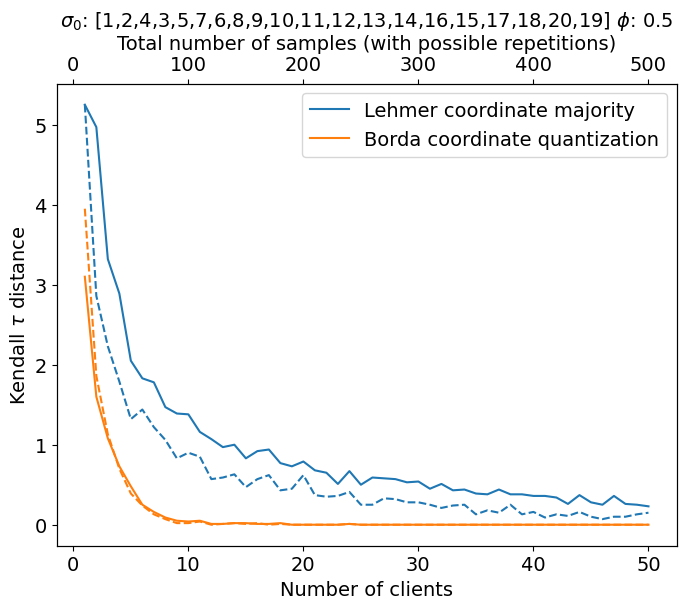} 
        \caption{$\sigma_0 \neq e, \phi=0.5$, $m_l=10$.} \label{}
    \end{subfigure}
    \begin{subfigure}[t]{0.49\textwidth}
        \centering
        \includegraphics[width=\textwidth]{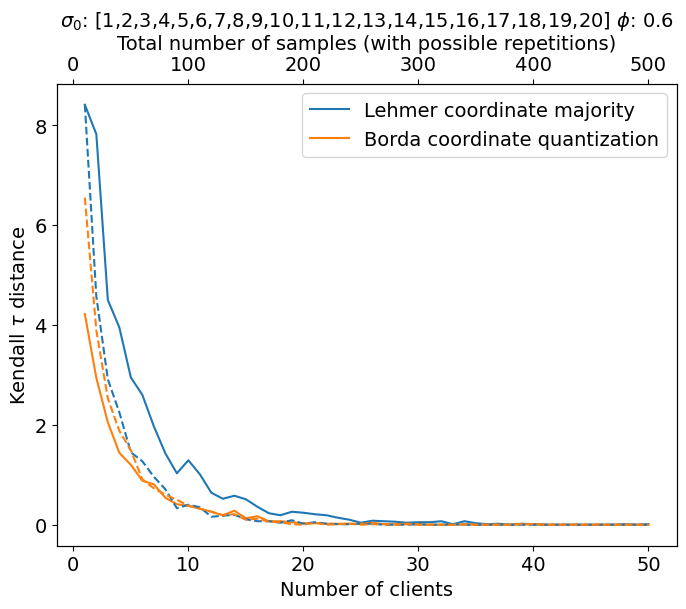} 
        \caption{$\sigma_0 = e, \phi=0.6$, $m_l=10$.} \label{}
    \end{subfigure}
    \hfill
    \begin{subfigure}[t]{0.49\textwidth}
        \centering
        \includegraphics[width=\textwidth]{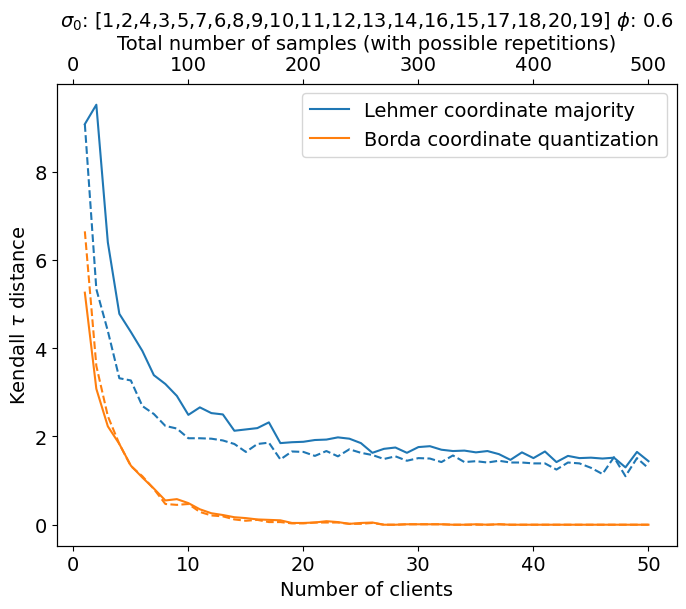} 
        \caption{$\sigma_0 \neq e, \phi=0.6$, $m_l=10$.} \label{}
    \end{subfigure}

    \caption{Comparison of the performance of various centralized and FRA methods (centralized results are depicted by dashed, while federated results are indicated by solid lines). Plots show the average Kendall $\tau$ distances between $\sigma_0$ and its estimate, for different values of $\phi$.}
\end{figure*}

\end{document}